\documentclass[manuscript,screen,nonacm]{acmart}

\AtBeginDocument{%
  }
    
\usepackage{adjustbox}
\usepackage[graphicx]{realboxes}
\usepackage{mathtools}
\usepackage{multirow}
\usepackage{color, colortbl}
\usepackage{paralist}
\usepackage{pgfplots}
\usepackage{subfigure}
\usepackage{threeparttable}
\usepackage{tikz}
\usetikzlibrary{positioning,shapes,arrows,shadows}
\pgfplotsset{compat=1.8}
\usepackage{pgfplotstable}
\usepackage{geometry} 

\usepackage{rotating}

\usepackage{makecell}
\usepackage{setspace}

\definecolor{Gray}{gray}{0.97}
\renewcommand{\vec}{\mathbf}

\def\hl{\setlength{\fboxsep}{1.0pt}\colorbox[rgb]{0.85,0.85,0.85}}

\newtheorem*{example*}{Example}

\usepackage{xpatch}

\makeatletter
\xpatchcmd{\ps@firstpagestyle}{Manuscript submitted to ACM}{}{\typeout{First patch succeeded}}{\typeout{first patch failed}}
\xpatchcmd{\ps@standardpagestyle}{Manuscript submitted to ACM}{}{\typeout{Second patch succeeded}}{\typeout{Second patch failed}}    \@ACM@manuscriptfalse
\makeatother

\renewcommand\footnotetextcopyrightpermission[1]{} 
\setcopyright{none}
\title[EDOLAB: A Platform for Evolutionary Dynamic Optimization Algorithms]{EDOLAB: An Open-Source Platform for Education and Experimentation with Evolutionary Dynamic Optimization Algorithms}

\author{Mai~Peng}
\email{pengmai@cug.edu.cn}
\authornote{These authors contributed equally to this work.}
\affiliation{%
  \institution{School of Automation, China University of Geosciences, Wuhan, Hubei Key Laboratory of Advanced Control and Intelligent Automation for Complex Systems, and Engineering Research Center of Intelligent Technology for Geo-Exploration, Ministry of Education}
  \country{China}
 \postcode{430074}
}

\author{Delaram~Yazdani}
\email{delaram.yazdani@yahoo.com}
\authornotemark[1]
\affiliation{%
  \institution{Liverpool Logistics, Offshore and Marine (LOOM) Research Institute, Faculty of Engineering and Technology, Liverpool John Moores University}
  \city{Liverpool}
  \country{United Kingdom}
 \postcode{L3 3AF}
}

\author{Zeneng~She}
\email{20s151103@stu.hit.edu.cn}
\authornotemark[1]
\affiliation{%
  \institution{School of Computer Science and Technology, Harbin Institute of Technology}
  \city{Shenzhen}
  \country{China}
 \postcode{518055}
}

\author{Danial~Yazdani}
\email{danial.yazdani@gmail.com}
\authornotemark[1]
\affiliation{%
  \institution{Faculty of Engineering \& Information Technology, University of Technology Sydney}
  \city{Ultimo}
  \country{Australia}
  \postcode{2007}
}

\author{Wenjian~Luo}
\email{ luowenjian@hit.edu.cn}
\authornotemark[1]
\affiliation{%
  \institution{Guangdong Provincial Key Laboratory of Novel Security Intelligence Technologies, School of Computer Science and Technology, Harbin Institute of Technology and Peng Cheng Laboratory}
  \city{Shenzhen}
  \country{China}
 \postcode{518055}
}

\author{Changhe~Li}
\email{changhe.lw@gmail.com}
\authornotemark[1]
\affiliation{
  \institution{School of Artificial Intelligence, Anhui University of Sciences \& Technology}
  \city{Hefei}
  \country{China}
 \postcode{230026}
}

\author{Juergen~Branke}
\email{Juergen.Branke@wbs.ac.uk}
\affiliation{%
  \institution{Information Systems Management and Analytics in Warwick Business School, University of Warwick}
  \city{Coventry}
  \country{United Kingdom}
  \postcode{CV4 7AL}
}

\author{Trung~Thanh~Nguyen}
\email{T.T.Nguyen@ljmu.ac.uk}
\affiliation{%
  \institution{The Liverpool Logistics, Offshore and Marine (LOOM) Research Institute, Faculty of Engineering and Technology, Liverpool John Moores University}
  \city{Liverpool}
  \country{United Kingdom}
  \postcode{L2 2ER}
}

\author{Amir~H.~Gandomi}
\email{Gandomi@uts.edu.au}
\authornote{Corresponding author}
\affiliation{%
  \institution{Faculty of Engineering \& Information Technology, University of Technology Sydney}
  \city{Ultimo}
  \country{Australia}
  \postcode{2007}
}
\affiliation{%
  \institution{University Research and Innovation Center (EKIK), Obuda University}
  \city{Budapest}
  \country{Hungary}
  \postcode{1034}
}

\author{Shengxiang Yang}
\email{syang@dmu.ac.uk}
\affiliation{%
  \institution{Institute of Artificial Intelligence (IAI), School of Computer Science and Informatics, De Montfort University}
  \city{Leicester}
  \country{United Kingdom}
  \postcode{LE1 9BH}
}

\author{Yaochu~Jin}
\email{jinyaochu@westlake.edu.cn}
\affiliation{%
  \institution{Department of Artificial Intelligence, School of Engineering, Westlake University}
  \city{Hangzhou}
  \country{China}
  \postcode{301130}
}

\author{Xin~Yao}
\email{xinyao@ln.edu.hk}
\authornotemark[2]
\affiliation{%
  \institution{School of Data Science, Lingnan University}
  \country{Hong Kong SAR}
}
\affiliation{%
  \institution{The Center of Excellence for Research in Computational Intelligence and Applications (CERCIA), School of Computer Science, University of Birmingham}
  \city{Birmingham}
  \country{United Kingdom}
  \postcode{B15 2TT}
}
\thanks{
This work was supported by the National Natural Science Foundation of China (Grant No. 62250710682), Shenzhen Fundamental Research Program (Grant No. JCYJ20220818102414030), Guangdong Provincial Key Laboratory of Novel Security Intelligence Technologies (Grant No. 2022B1212010005), the National  Natural Science Foundation of China (Grant No. 62076226),  the Fundamental Research Funds for the Central Universities China University of Geosciences (Wuhan) (Grant No. CUGGC02), and the Australian Government through the Australian Research Council under Project DE210101808. }   

\setcopyright{none}
\renewcommand\footnotetextcopyrightpermission[1]{}
\begin{document}
 \newpage
\begin{abstract}
\textit{\textbf{Abstract---}}
Many real-world optimization problems exhibit dynamic characteristics, posing significant challenges for traditional optimization techniques. 
Evolutionary Dynamic Optimization Algorithms (EDOAs) are designed to address these challenges effectively. 
However, in existing literature, the reported results for a given EDOA can vary significantly. This inconsistency often arises because the source codes for many EDOAs, which are typically complex, have not been made publicly available, leading to error-prone re-implementations.
To support researchers in conducting experiments and comparing their algorithms with various EDOAs, we have developed an open-source MATLAB platform called the \textit{E}volutionary \textit{D}ynamic \textit{O}ptimization \textit{LAB}oratory (EDOLAB). 
This platform not only facilitates research but also includes an educational module designed for instructional purposes. 
The education module allows users to observe: \begin{inparaenum} [a)] \item a 2-dimensional problem space and its morphological changes following each environmental change, \item the behaviors of individuals over time, and \item how the EDOA responds to environmental changes and tracks the moving optimum. \end{inparaenum} 
The current version of EDOLAB features 25 EDOAs and four fully parametric benchmark generators. 
 \end{abstract}


\maketitle



\section{Introduction}
\label{sec:Introduction}

Many real-world optimization problems are dynamic in nature~\cite{nguyen2011thesis}, meaning that the characteristics of their search spaces change over time~\cite{yazdani2021DOPsurveyPartA,raquel2013dynamic}. 
These environmental changes in dynamic optimization problems (DOPs) introduce uncertainties that must be accounted for by the optimization algorithm~\cite{jin2005evolutionary}. 
To effectively solve DOPs, it is important that the optimization algorithm not only locates an optimal solution efficiently but also continues to track it as the environment changes. 
This capability is referred to as tracking the moving optimum (TMO)~\cite{nguyen2012evolutionary}.

Evolutionary algorithms and swarm intelligence methods are popular and effective optimization tools, originally designed for solving static optimization problems. 
Applying these tools directly to TMO in DOPs proves ineffective because they fail to account for environmental changes. 
These changes in DOPs present several challenges, including: 
\begin{inparaenum} [a)] 
\item outdated stored fitness values (also known as objective function values), 
\item local and global diversity loss, and 
\item a limited number of objective function evaluations that can be conducted between consecutive environmental changes (i.e., within each environment)~\cite{yazdani2020adaptive}. 
\end{inparaenum} 
To address the challenges of optimizing in dynamic environments and performing TMO in DOPs, evolutionary algorithms and swarm intelligence methods are typically augmented with additional components, forming evolutionary dynamic optimization algorithms (EDOAs). 
These components may include local and global diversity control, explicit archives, change detection and reaction mechanisms, population clustering and management, exclusion, convergence detection, and computational resource allocation~\cite{yazdani2021DOPsurveyPartA}. 
Consequently, EDOAs often become complex algorithms.

{\color{black}The complexity of EDOAs, especially state-of-the-art ones, makes them challenging to re-implement accurately. 
Even minor changes or mistakes can significantly impact their performance. For instance, altering the order in which mechanisms are executed or the timing of their activation could significantly change the behavior of the algorithms.}

The lack of publicly available source codes for many EDOAs has posed a significant challenge for researchers attempting to reproduce results for experimentation and comparison. 
In addition to the complexities inherent in EDOAs, accurately calculating performance indicators and generating dynamic benchmarks is also a complex process. 
Upon reviewing some of the limited available source codes, we discovered that certain performance indicators, such as offline error~\cite{branke2003designing}, are often calculated incorrectly. 
Moreover, in other cases, parameters of the widely used Moving Peaks Benchmark (MPB)~\cite{branke1999memory}, including random number generators and initial peak values, are configured in ways that lead to unfair comparisons.
Additionally, there is currently no comprehensive software platform available for evaluating the performance of EDOAs and for identifying both their strengths and weaknesses in solving DOP instances with various morphological and dynamic characteristics~\cite{herring2022reproducibility}.

To address the pressing need for such software, we have developed an open-source MATLAB platform for EDOAs, known as the \textit{E}volutionary \textit{D}ynamic \textit{O}ptimization \textit{LAB}oratory (EDOLAB).
The current version of EDOLAB is primarily focused on single-objective, unconstrained, continuous DOPs. 
However, the EDOAs designed for this class of DOPs have been demonstrated to be easily extendable to address other significant classes of DOPs, including robust optimization over time (ROOT)~\cite{yazdani2023robust,yazdani2022robust}, constrained DOPs~\cite{nguyen2012continuous,bu2016continuous}, and large-scale DOPs~\cite{luo2017hybrid,yazdani2019scaling}.

Moreover, although the structures of EDOAs designed for single-objective DOPs differ from those tailored for finding Pareto optimal solutions (POS) in each environment within multi-objective DOPs~\cite{jiang2022evolutionary}, they remain effective for addressing many multi-objective DOPs. 
In fact, in many multi-objective DOPs, a \emph{single} solution is deployed in each environment, chosen by a decision maker based on both user preferences and problem-specific characteristics. 
Therefore, finding the POS for each environment and selecting a solution for deployment may not always be the most effective approach, particularly in problems with high-frequency environmental changes.
{\color{black}For example, given a real-world multi-objective DOP where the environment changes every few seconds, it can be challenging for a user to select a solution from the POS for each environment. 
To address such multi-objective DOPs, the problem can be transformed into a single-objective DOP by combining all objectives according to the decision maker's preferences. 
These preferences are both user-driven, based on the decision maker's goals and values, and problem-dependent, considering the specific nature and constraints of the problem at hand~\cite{marler2010weighted,kaddani2017weighted}.
Consequently, in the resulting single-objective problem, a single-objective EDOA may be more suitable, focusing on finding an optimal solution for deployment in each environment based on these combined preferences.}

In the following, we describe the major contributions and features of EDOLAB:
\paragraph{\textit{Comprehensive library}}
The current release of EDOLAB includes 25 EDOAs, as listed in Table~\ref{tab:algorithms}, each with distinct characteristics such as varying structures, optimizers, population clustering and sub-population management methods, diversity control components, and computational resource allocation strategies.
EDOLAB also includes {\color{black}four dynamic benchmark generators: MPB~\cite{branke1999memory},  free peaks (FPs)~\cite{li2018open},  Generalized Dynamic Benchmark Generator (GDBG)~\cite{li2008GDBG}, and Generalized MPB (GMPB)~\cite{yazdani2020benchmarking,yazdani2021generalized}.}
These benchmark generators are fully parametric and capable of producing dynamic problem instances with varying morphological and dynamical characteristics.
To evaluate the efficiency of EDOAs in solving DOPs and facilitate comparisons, EDOLAB incorporates the two most commonly used performance indicators: offline error~\cite{branke2003designing} and the average error before environmental changes~\cite{trojanowski1999searching}.

\color{black}
\paragraph{\textit{Easy to use}}
EDOLAB is developed in MATLAB, a programming language that offers a vast collection of high-level mathematical functions for operating on arrays and matrices, along with various random number generators. 
These features make MATLAB an ideal choice for implementing EDOAs and dynamic benchmarks. 
The source codes of EDOLAB, particularly for the EDOAs and benchmarks, are designed to be easy to understand, trace, and modify, thanks to MATLAB's capabilities.

Based on our investigations, MATLAB has been one of the most frequently used programming languages in the field of evolutionary dynamic optimization due to its strengths and overall ease of use, including its straightforward syntax and readability.
Moreover, MATLAB is highly popular not only in this field but also in other active areas such as evolutionary multi-objective optimization. This popularity is reflected in the widespread use of platforms like PlatEMO~\cite{tian2017platemo}, which is implemented in MATLAB.

We have also structured and modularized the platform to ensure the clarity and readability of the source code. Informative parameter names, clear distinctions between components, and comprehensive comments make the source code easy to trace and modify. Additionally, EDOLAB features a graphical user interface (GUI) to enhance user-friendliness, making it accessible even to beginners. The GUI enables users to effortlessly select an EDOA, configure a problem instance, and run experiments with minimal effort.

\color{black}

\paragraph{\textit{Flexible and Comprehensive Research Capabilities}}
EDOLAB offers researchers significant flexibility in conducting empirical studies, with the option to operate the platform either with or without the GUI. 
Its extensive library allows for the thorough investigation and comparison of various EDOAs, each designed with different structures, optimizers, and components to address dynamic optimization challenges across a wide range of problem instances.
Additionally, EDOLAB is particularly valuable for researchers developing new EDOAs or improving existing algorithms.
Since the platform's source code is open-source and modularly designed, it becomes easy to incorporate new mechanisms---such as population control, diversity control, or other innovative components---into the existing algorithms. 
The platform includes four dynamic benchmark generators, capable of producing a broad range of problem instances with varying levels of difficulty and characteristics. 
This, coupled with a collection of comparison EDOAs, performance indicators, and visualization plots, makes EDOLAB an invaluable tool for evaluating algorithms and generating results for scientific reports and articles.

\color{black}
\paragraph{\textit{Educational Support}}
EDOLAB includes an educational module, which is particularly useful for new researchers, such as PhD students, to enter the field and contribute more efficiently. 
This module visualizes the 2-dimensional problem space (i.e., environment) and dynamically illustrates how the morphology of the search space evolves after each environmental change. 
Users can observe how individuals relocate over time, providing valuable insights into how EDOAs adapt to environmental changes. 
By highlighting the similarity factors between successive environments, this module enables users to grasp the significance of knowledge transfer from the previous to the current environment, accelerating their understanding of complex dynamic optimization behaviors.
\color{black}

\paragraph{\textit{Extensibility}}
EDOLAB is designed for easy extensibility, allowing researchers to expand its library by adding new EDOAs, benchmark problems, and performance indicators. 
With EDOLAB's open-source code, researchers can also modify EDOA frameworks, incorporate their own components, and explore their effectiveness. 
For instance, if a researcher develops a new exclusion, change reaction, or convergence detection component, they can simply replace the existing code with the new one and assess its impact on the overall performance of an EDOA.
Additionally, researchers can extend EDOAs to address other classes of DOPs by incorporating specific components necessary for those problems, such as constraint-handling mechanisms for constrained DOPs~\cite{nguyen2012continuous} or decision-making components for altering or maintaining deployed solutions in ROOT~\cite{yazdani2017new,yazdani2018changing}.

\color{black}
\paragraph{\textit{Innovative modular design}}
One of the key contributions of EDOLAB is its innovative design, which unifies a diverse set of EDOAs through a modular structure. This design allows different algorithms, each with varying mechanisms and structures, to be integrated into a single, cohesive platform. By utilizing a novel approach to component modularization, EDOLAB enables these algorithms to share a common framework while maintaining their unique characteristics. In addition, the modular design supports the extensibility of the platform, allowing new algorithms, benchmarks, and performance indicators to be incorporated with minimal effort.

\paragraph{\textit{Open-source availability and accessibility}}
A significant contribution of EDOLAB is its open-source availability. 
The platform is publicly accessible through GitHub, allowing researchers to utilize and extend its functionality for experimental and comparative studies. 
By making EDOLAB open-source, we aim to provide a valuable resource for researchers working on evolutionary dynamic optimization. 
The platform can be accessed from [\url{https://github.com/EDOLAB-platform/EDOLAB-MATLAB}].
\color{black}

The remainder of this paper is organized as follows: Section~\ref{sec:contains} provides definitions of DOPs, benchmark generators, performance indicators, and the EDOAs included in EDOLAB.
Section~\ref{sec:Overview} offers an overview of the platform's structure and architecture. The technical aspects of EDOLAB, including its software architecture, source code structure, usage methods, GUI, and extension capabilities, are detailed in the user manual, which is provided as a separate document. 
Finally, Section~\ref{sec:conclusion} provides the concluding remarks of the paper.

\section{EDOLAB's library}
\label{sec:contains}

We begin this section by defining the sub-field of dynamic optimization considered in the current version of EDOLAB, which is single-objective, unconstrained, continuous DOPs. 
Note that all EDOAs, benchmark generators, and performance indicators in the EDOLAB library are specifically developed for maximization problems. 
We then describe the EDOLAB library, which includes four dynamic benchmark generators, two performance indicators, and 25 EDOAs.

A single-objective, unconstrained, continuous DOP can be defined as: 
\begin{align}
\label{eq:DOP1}
\mathrm{Maximize:} & \;\; f^{(t)}(\vec{x})=f\left(\vec{x},\vec{\alpha}^{(t)}\right),\;\vec{x}=\{x_1,x_2,\cdots,x_d\},
\end{align}
where $\vec{x}$ is a solution in the $d$-dimensional search space, $f$ is the time-varying objective function, $t$ is the time index, and $\vec{\alpha}$ is a set of time-varying environmental parameters.
Most research in this field focuses on DOPs where environmental changes occur only at discrete time steps, which is characteristic of many real-world DOPs~\cite{nguyen2011thesis}. 
For a problem with $T$ environments, there is a sequence of $T$ stationary search spaces. 
Consequently, a DOP, with $T$ environmental states (i.e., $T-1$ environmental changes), can be reformulated as:
{\color{black}\begin{align}
\label{eq:DOP2}
 \mathrm{Maximize:} &   f(\vec{x}) = \left\{ f(\vec{x},\vec{\alpha}^{(t)}) \right\}_{t=1}^{T} = \left\{ f(\vec{x},\vec{\alpha}^{(1)}),f(\vec{x},\vec{\alpha}^{(2)}), \dots ,f(\vec{x},\vec{\alpha}^{(T)}) \right\}.
\end{align}}
It is generally assumed that there is a degree of morphological similarity between successive environments, a characteristic commonly observed in many real-world DOPs~\cite{yazdani2018thesis,nguyen2011thesis,branke2012evolutionary}.

\subsection{Benchmark generators}
\label{sec:sec:benchmarks}

MPB~\cite{branke1999memory} is the most widely used dynamic benchmark generator in the field~\cite{yazdani2020benchmarking,yazdani2021DOPsurveyPartB}. 
The landscapes generated by the standard version of MPB are relatively straightforward to optimize, as they consist of a series of conical promising regions (i.e., peaks) that are regular, unimodal, symmetric, fully separable~\cite{yazdani2019scaling}, and well-conditioned. 
Despite its simplicity, MPB is an essential component of EDOLAB due to its significant value for educational purposes.
MPB's straightforward nature makes it easier for users to observe the behavior of EDOAs over time and investigate the effectiveness of various components, such as promising region coverage, change reaction, and diversity control. 
In the standard MPB, the height, width, and center of each promising region (represented as a cone) change over time. 
To enhance the realism of MPB in EDOLAB, we have made a few modifications. 
First, we removed the option that allowed all promising regions to be initialized with identical height and width. 
In EDOLAB, the attributes of all promising regions are initialized randomly within predefined ranges. 
Second, we eliminated the option for correlated movements of promising regions, which could result in linear relocation directions of the promising region centers. 
Instead, in EDOLAB, the directions of shifts are randomized, providing a stochastic and more challenging dynamic environment.

{\color{black}GDBG~\cite{li2008GDBG} is the second most commonly used dynamic benchmark in the field. 
GDBG is created by introducing dynamics to composition benchmark functions~\cite{suganthan2005problem,liang2005novel}, which are commonly employed in static global optimization. 
The problem instances generated by GDBG are generally more complex and challenging than those produced by MPB, as they involve landscapes with irregular, multimodal, and partially separable components. 
Despite its lack of fully controllable characteristics, GDBG has been widely used in research and can still provide valuable insights into the performance of EDOAs. 
To create a more comprehensive platform, we have included GDBG in EDOLAB, allowing researchers to leverage its complexity for a broader range of experimental evaluations.}

FPs benchmark~\cite{li2018open} is the third benchmark generator included in EDOLAB. 
The landscapes generated by FPs are divided into several hypercubes using a k-d tree~\cite{bentley1979data}, with each hypercube containing one promising region. 
As a result, the basin of attraction for each promising region is determined by the hypercube in which it lies. 
After each environmental change, the shape of each promising region is randomly selected from eight different unimodal functions. 
Additionally, the location and size of each hypercube change over time, which alter the basin of attraction of the promising region. 
The center position of each promising region also shifts within the hypercube. 
Several transformations, such as symmetry breaking and condition number increasing, are applied in FPs, though these transformations remain fixed over time. 
FPs is also suitable for educational purposes, as its promising regions are clearly defined by the hypercubes.

GMPB~\cite{yazdani2020benchmarking,yazdani2021generalized} is the final benchmark generator included in EDOLAB.
GMPB is a complex, fully configurable benchmark generator. 
The landscapes generated by GMPB are constructed by assembling several promising regions with a variety of controllable characteristics, ranging from unimodal to highly multimodal, symmetric to highly asymmetric, smooth to highly irregular, and varying degrees of variable interaction and ill-conditioning. 
All of these characteristics can change over time. 
With its high degree of configurability, GMPB allows users to examine the performance of proposed or existing EDOAs across a wide range of problem instances with different characteristics and difficulty levels. 
Consequently, GMPB is well-suited for experimentation and for investigating and comparing the performance of different EDOAs. 
However, due to the complexity of the search spaces generated by GMPB, it is not the ideal choice for educational purposes.

Figure~\ref{fig:benchmarks} provides examples of the landscapes generated by the four benchmarks included in EDOLAB: MPB, GDBG, FPs, and GMPB. 
These contour plots illustrate the morphological characteristics of each benchmark's problem space. However, it is important to note that these are just examples, and the characteristics shown in these figures cannot be generalized to all possible landscapes that can be generated by these benchmarks.

\begin{figure*}[!t]
\centering
\begin{tabular}{cc}
    \subfigure[{\scriptsize Example of a landscape generated by the MPB benchmark.}]{\includegraphics[width=0.45\linewidth]{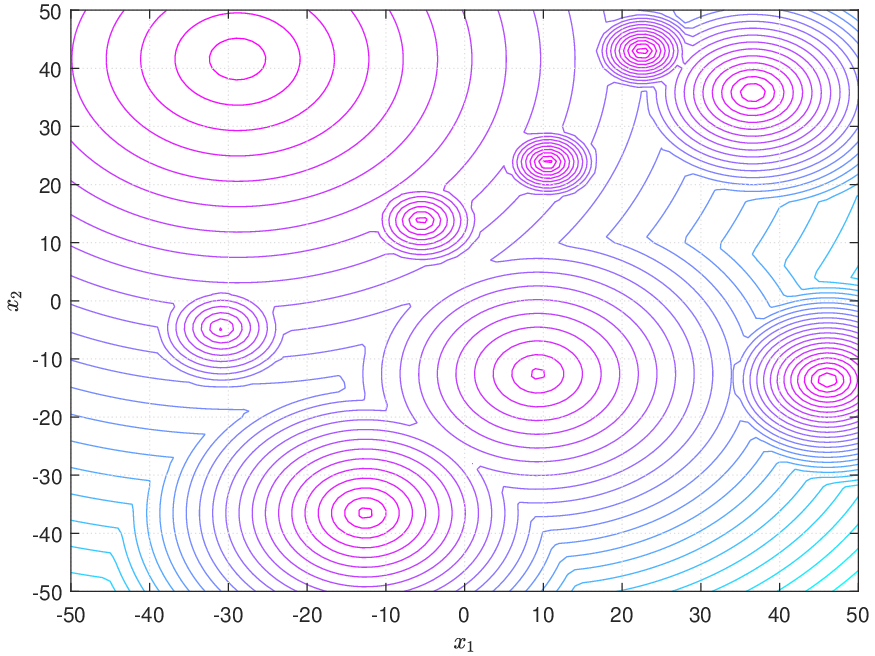}\label{fig:MPB}}
&
     \subfigure[{\scriptsize Example of a landscape generated by the GDBG benchmark.}]{\includegraphics[width=0.45\linewidth]{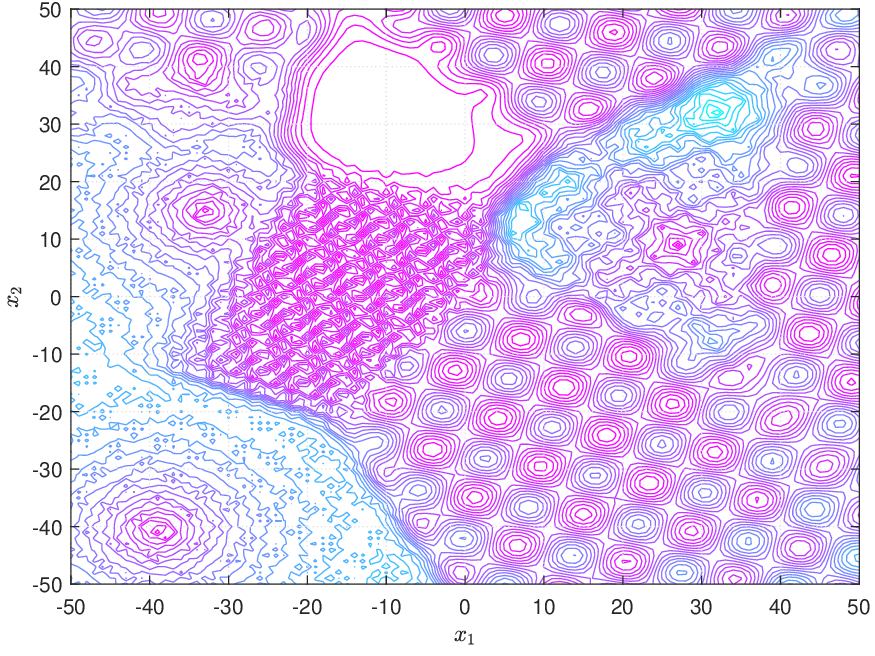}\label{fig:GDBG}}
\\
    \subfigure[{\scriptsize Example of a landscape generated by the FPs benchmark.}]{\includegraphics[width=0.45\linewidth]{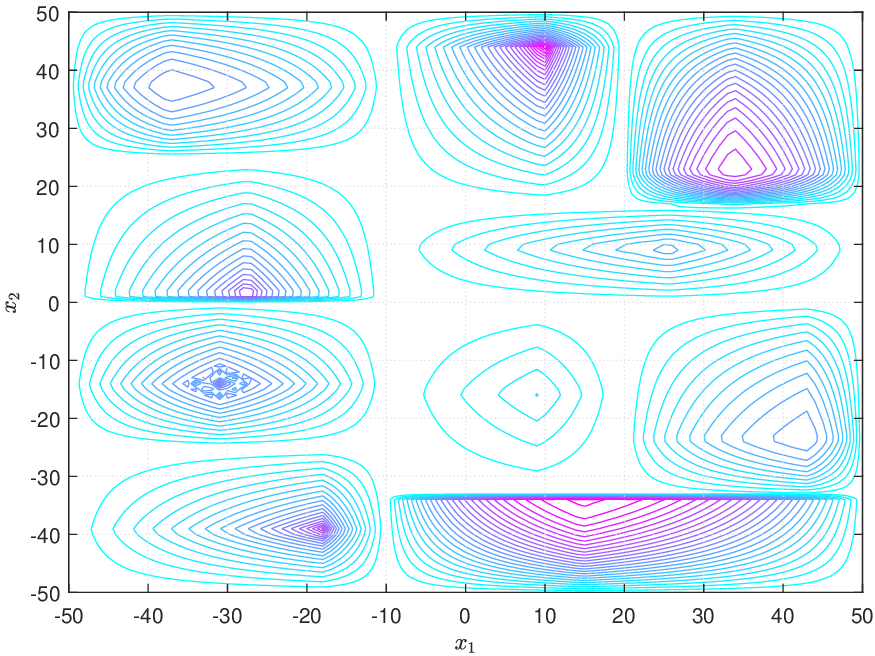}\label{fig:FPs}}
&
     \subfigure[{\scriptsize Example of a landscape generated by the GMPB benchmark.}]{\includegraphics[width=0.45\linewidth]{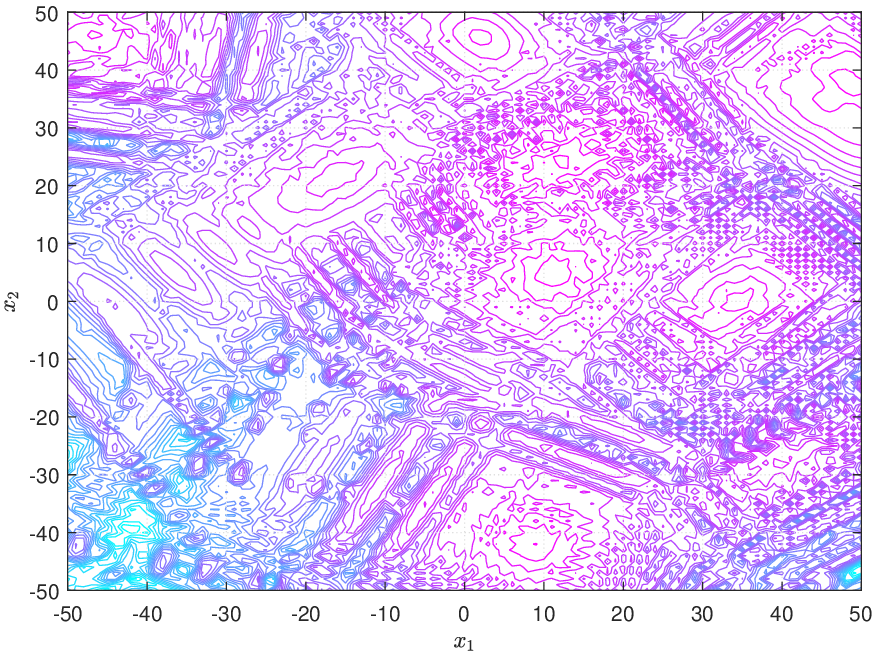}\label{fig:GMPB}}
\end{tabular}
\caption{Examples of two-dimensional landscapes generated by the four benchmarks: (a) MPB, (b) GDBG, (c) FPs, and (d) GMPB. These are representative examples, and the characteristics shown cannot be generalized to all possible landscapes generated by these benchmarks.}
\label{fig:benchmarks}

\label{fig:benchmarks}
\end{figure*}

\color{black}
The benchmarks included in EDOLAB provide a flexible and robust framework for evaluating the performance of EDOAs. 
Their parametric nature allows researchers to generate a wide variety of problem instances with different morphological and dynamical characteristics, making them invaluable for algorithm evaluation and educational purposes. 
These benchmarks help researchers systematically understand how algorithms behave under controlled scenarios, revealing their strengths and weaknesses, which is crucial for guiding future improvements. 
Additionally, they serve as a foundation for new researchers to study and experiment with algorithmic concepts, making them highly suitable for educational use.
In the user manual, we have provided commonly used parameter settings for generating problem instances from the four benchmark generators included in EDOLAB. 
These settings generate 12 instances per benchmark generator, covering a range of difficulties and characteristics that are commonly used in the literature.

At this stage, we do not include real-world problems in the platform for two primary reasons. First, many real-world problems are inflexible and do not allow the generation of diverse problem instances with varying characteristics, limiting their usefulness for the comprehensive evaluation of algorithms. 
Second, for many real-world problems, we do not fully understand their morphological characteristics, dynamic behavior, or the specific challenges they present. 
This lack of transparency makes them less suitable for testing and refining algorithms.

While there are real-world problem domains, such as dynamic facility location problems, that offer the flexibility to generate instances with controllable characteristics, the algorithms currently available in the field, including those in EDOLAB, are not yet capable of addressing the complex challenges posed by such problems~\cite{yazdani2024clustering}. 
It is not simply a matter of poor performance; these algorithms are unable to function effectively in these scenarios, underscoring the need for further advancements in the field before tackling such real-world challenges. 
As a result, we have chosen not to include these problems or benchmarks that simulate their behavior.
\color{black}

\subsection{Performance indicators}
\label{sec:sec:indicator}

Since the global optimum is known for all the benchmark generators used in EDOLAB, error-based performance indicators are well-suited for evaluating the performance of EDOAs~\cite{yazdani2021DOPsurveyPartB}. 
In EDOLAB, we employ two key error-based performance indicators: offline error and average error before environmental changes.

\subsubsection{Offline error}

Offline error measures the mean error of the best-found solution across all objective function evaluations. 
It is calculated using the following equation~\cite{branke2003designing}:
\begin{align}
\label{eq:Eo}
E_\mathrm{O} = \frac{1} { \sum_{t=1}^T \upsilon^{(t)} } \sum_{t=1}^T \sum_{\vartheta=1}^{\upsilon^{(t)}} \left(f^{(t)}\left( \vec{\mathrm{g}}^{\star(t)} \right) - f^{(t)}\left(\vec{\mathrm{g}}^{*\left(\vartheta+\sum_{k=1}^{(t-1)} \upsilon^{(k)}  \right)}\right)\right),
\end{align}
where $E_\mathrm{O}$ represents the offline error, $\vec{\mathrm{g}}^{\star(t)}$ is the global optimum at the $t$-th environment, $\upsilon^{(t)}$ is the number of objective function evaluations in the $t$-th environment, $T$ is the number of environments, $\vartheta$ is the function evaluation counter for each environment, and $\vec{\mathrm{g}}^{*\left(\vartheta+\sum_{k=1}^{(t-1)} \upsilon^{(k)} \right)}$ is the best-found solution at the $\vartheta$-th function evaluation in the $t$-th environment.
{\color{black} The offline error measures the overall performance of an algorithm across all function evaluations and captures the convergence speed of an EDOA following environmental changes. This metric serves as an effective indicator for assessing how quickly (based on the number of function evaluations) an algorithm adapts to changes in the environment and converges towards the optimal solution.}

\subsubsection{Average error before environmental changes}

The second performance indicator, average error before environmental changes ($E_\mathrm{BBC}$)~\cite{trojanowski1999searching}, focuses on the error of the best-found solution just before each environmental change. It is calculated as follows:
\begin{align}
\label{eq:Eb}
E_\mathrm{BBC} = \frac{1}{T} \sum_{t=1}^T  \left(f^{(t)}\left( \vec{\mathrm{g}}^{\star(t)} \right) - f^{(t)}\left(\vec{\mathrm{g}}^{\mathrm{end}(t)}\right)\right),
\end{align}
where $\vec{\mathrm{g}}^{\mathrm{end}(t)}$ is the best found solution at the end of the $t$-th environment. 
{\color{black} Since $E_\mathrm{BBC}$ focuses solely on the best-found solution at the end of each environment, it does not capture the convergence speed or performance across all function evaluations within an environment.}

These two performance indicators---offline error and average error before environmental changes---are the most widely used in the field, with over 85\% of studies relying on them~\cite{yazdani2021DOPsurveyPartB}. 
In EDOLAB, we chose not to include the offline performance~\cite{branke1999memory}, a third commonly used indicator, as it provides no additional insights beyond what is captured by the offline error.
Furthermore, we do not incorporate distance-to-optimum-based performance indicators~\cite{duhain2012particle} in EDOLAB. 
These indicators evaluate performance based on the proximity of the closest found solution to the global optimum, whereas nearly all EDOAs are designed to optimize based on fitness values. 
As such, using indicators that do not account for the quality of the found solutions (i.e., fitness value) may lead to inaccurate assessments of EDOA behavior~\cite{yazdani2021DOPsurveyPartB}.

\color{black}
In addition to these two performance indicators, EDOLAB provides two types of plots that assist in analyzing the algorithm's performance across all function evaluations: \emph{current error plots} and \emph{offline error over time plots}.
\begin{itemize}
    \item {Current error plots} display the convergence behavior of the algorithm within each environment, showing how the error evolves over function evaluations. These plots reflect how quickly an algorithm reacts to environmental changes and converges toward the optimum during each environment. At each function evaluation, the current error plot presents the error of the best-found solution.
    \item {Offline error over time plots} illustrate the offline error value across all function evaluations. This means that for each function evaluation, the plot shows the average error of the best-found solutions across all previous function evaluations.
\end{itemize}
\color{black}

\subsection{EDOAs}
\label{sec:EDOAs}

\color{black}
\subsubsection{\textbf{Overview of Common Frameworks in Evolutionary Dynamic Optimization}}
\label{sec:sec:EDOAoverview}

Below, we provide an overview of the common frameworks and components used in EDOAs. 
For a more detailed analysis, comprehensive descriptions, and in-depth taxonomy of algorithms and their components, we refer readers to several key surveys on the subject~\cite{mavrovouniotis2017survey,nguyen2012evolutionary,yazdani2024survey,yazdani2021DOPsurveyPartA}.

EDOAs are complex algorithms composed of multiple components designed to address challenges in DOPs. 
These components work together to enhance the algorithm's ability to track the moving optimum as the environment changes. 
The ongoing development of new and improved components is a key area of research, with the goal of further advancing the performance of EDOAs. 
Below, we describe the major components commonly found in EDOA frameworks~\cite{yazdani2021DOPsurveyPartA}:

\begin{itemize}
\item \textbf{Population Management:} These components are responsible for controlling the creation, organization, and activities of subpopulations in algorithms that utilize more than one population, such as bi-population and, more commonly, multi-population methods~\cite{li2015multipopulation}. 
They dictate how subpopulations are formed, how they share information with each other, and how computational resources are allocated across the subpopulations to ensure an effective balance between exploration and exploitation.
\item \textbf{Explicit Memory:} EDOAs often rely on historical knowledge to enhance their optimization process over time and after environmental changes. 
One method for storing and retrieving this information is through explicit memory~\cite{branke1999memory}, which keeps track of past optima. In some algorithms, explicit memory is used to accelerate the change reaction, while in others, it is employed to manage subpopulations and prevent over-exploitation of promising regions.
\item \textbf{Diversity Control:} In evolutionary algorithms, \textit{diversity loss} is a significant challenge when optimizing in dynamic environments~\cite{branke2012evolutionary}.  
Diversity control components help address this challenge. 
There are two main types of diversity control components. 
The first type, \textit{local diversity control}, addresses local diversity loss by facilitating the tracking of local optima and enhancing exploitation after environmental changes. 
The second type, \textit{global diversity control}, aims to maintain the capability of exploration. 
Diversity control components are further classified into those that maintain diversity throughout the optimization process and those that inject diversity into subpopulations or the overall population when certain conditions trigger them.
\item \textbf{Convergence Detection:} While not directly addressing any specific challenge of optimization in dynamic environments, convergence detection is essential for triggering other components, such as those related to diversity control or population management. It helps identify when a population or subpopulation has converged on a promising region.
\item \textbf{Optimizer:} One of the core components of an EDOA is the optimizer itself, which is often an algorithm originally designed for static optimization. However, when integrated with other components such as population management and diversity control, these optimizers become capable of handling dynamic environments by adapting to environmental changes and maintaining effective search performance. 
\end{itemize}

A significant line of research in the field involves the development of new versions or improvements of these core components. 
In EDOLAB, we have designed the platform with modularity in mind, ensuring that each of the components is clearly separated and well-structured within the codebase. 
This modularization allows researchers to easily locate, modify, or replace individual components---such as the population management or diversity control mechanisms---without requiring a full reimplementation of the algorithm.
This design facilitates experimentation and enables users to evaluate the performance of existing algorithms when paired with newly developed components.

Based on the components and strategies used in the framework of EDOAs, we can categorize them into the following classes~\cite{yazdani2024survey}:

\begin{itemize}
    \item \textbf{Single-Population Algorithms}: These algorithms employ a single population throughout the optimization process~\cite{eberhart2001tracking}. 
    Their primary limitation is the lack of flexibility in exploring different regions of the search space, making them less effective for complex dynamic problems.
    
    \item \textbf{Bi-Population Algorithms}: Bi-population algorithms divide the population into two groups, where one typically focuses on exploration and the other on exploitation~\cite{branke1999memory}. This basic division offers improved performance compared to single-population algorithms by balancing the exploration of new regions and the exploitation of the best known promising region.
    
    \item \textbf{Multi-Population Algorithms}: The most advanced and commonly used class, multi-population EDOAs employ several subpopulations to explore various regions of the search space~\cite{blackwell2004multiswarm}. 
These algorithms tend to offer the best performance for solving dynamic optimization problems because they can simultaneously explore and exploit multiple regions. 
Multi-population algorithms are highly flexible and can incorporate a variety of other components such as diversity control and convergence detection.
 In terms of population management components, these algorithms can be further classified based on the following aspects:

    \begin{itemize}
        \item \textbf{Fixed vs. Adaptive Subpopulations}: In multi-population methods, subpopulations can either have a fixed or adaptive structure. Algorithms with a fixed number of subpopulations maintain the same structure throughout the optimization process~\cite{blackwell2006multiswarm}, while those with adaptive subpopulations dynamically adjust the number of subpopulations based on the number of promising regions discovered~\cite{blackwell2007particle}. Adaptive algorithms rely on mechanisms for creating and eliminating subpopulations in response to environmental changes~\cite{yazdani2020adaptive}.
        
               \item \textbf{Clustering-Based Subpopulation Formation}: Subpopulations in multi-population EDOAs can be generated through clustering methods. These clustering strategies may use the positions and fitness values of individuals~\cite{yang2010clustering} or simply group individuals based on indices~\cite{blackwell2006multiswarm}. Clustering based on positions can be more effective, as it takes the spatial distribution of individuals into account when forming subpopulations~\cite{yazdani2021DOPsurveyPartA}. However, this approach is more computationally expensive as it requires frequent re-clustering of individuals.
        
  \item \textbf{Homogeneous vs. Heterogeneous Subpopulations}: Subpopulations in multi-population methods can be either homogeneous or heterogeneous. In homogeneous subpopulations, each subpopulation uses the same optimizer, structure, and parameter settings~\cite{mendes2006DynDE}. On the other hand, heterogeneous subpopulations~\cite{li2008fast}  have varying structures, parameter settings, or even optimizers, making them more flexible and potentially more effective at assigning different roles and covering diverse regions of the search space.

    \end{itemize}
\end{itemize}

\color{black}

\subsubsection{\textbf{EDOAs Included in EDOLAB}}
\label{sec:sec:CoveredEDOs}

\color{black}
The current release of EDOLAB includes 25 EDOAs, listed in Table~\ref{tab:algorithms}.  
As seen in the table, Particle Swarm Optimization (PSO)~\cite{bonyadi2017particle} and Differential Evolution (DE)~\cite{das2010differential} are the most commonly used optimizers in the algorithms included in EDOLAB.
This choice is in line with the distribution of optimizers in the literature, where these two are the most frequently employed in the field of evolutionary dynamic optimization~\cite{yazdani2021DOPsurveyPartB}.
Note that while Covariance Matrix Adaptation Evolution Strategy (CMA-ES)~\cite{CMA-ES} has proven to be a powerful and widely used optimizer in other fields of optimization, its application to continuous single-objective, unconstrained DOPs, which is the focus of the current version of EDOLAB, has been limited. 
Consequently, CMA-ES is only represented by a single algorithm in EDOLAB, which mirrors its relatively low adoption in the field of evolutionary dynamic optimization. 

Additionally, multi-population algorithms are heavily represented in EDOLAB, as they are widely recognized as the most effective class of algorithms for DOPs~\cite{yazdani2021DOPsurveyPartA,li2015multipopulation,yazdani2024survey}.
As discussed in Section~\ref{sec:sec:EDOAoverview}, different versions of multi-population EDOAs have been proposed in the literature, each employing a variety of techniques to manage subpopulations. 
In EDOLAB, we have included algorithms with:
\begin{itemize}
    \item Fixed and adaptive numbers of subpopulations,  
    \item Fixed and adaptive population sizes,  
    \item Various clustering methods for forming subpopulations, including clustering based on positions and fitness,  
    \item Homogeneous and heterogeneous subpopulation structures, where subpopulations may either share similar characteristics or differ in terms of parameter settings, size, or optimization components.
\end{itemize}
These variations provide a broad spectrum of strategies for addressing the complexities of DOPs.  
Overall, the distribution of EDOAs in EDOLAB is consistent with the trends observed in the literature, ensuring that the platform accurately represents the current state of the field.
\color{black}

\begin{table}[!t]
\small
\centering
  \caption{EDOAs included in the initial release of EDOLAB.} 
  \label{tab:algorithms}
\Rotatebox{90}{
    \begin{tabular}{l@{\ \ }p{0.16\textwidth}p{0.12\textwidth}c@{\ \ }c@{\ \ }c@{\ \ }p{0.1\textwidth}c@{\ \ }}
    \toprule
     \multirow{2}{*}{EDOA}         & \multirow{2}{*}{ Reference}        & Optimization & Population  & Number   of             & Population & Population & Sub-population\\
                                                                       &                                                                  &   component   &  structure     &  sub-populations &size &  clustering & heterogeneity\\
    \midrule 
\rowcolor{lightgray} ACF$_\mathrm{PSO}$& \cite{yazdani2020adaptive}&  PSO & Multi-population & Adaptive & Adaptive & By index & Homogeneous\\
AMP$_\mathrm{DE}$ &\cite{li2016adaptive}&DE& Multi-population & Adaptive & Adaptive & By position & Heterogeneous\\   
\rowcolor{lightgray}AMP$_\mathrm{PSO}$ &\cite{li2016adaptive}&PSO & Multi-population & Adaptive & Adaptive & By position  & Heterogeneous\\     
AmQSO      & \cite{blackwell2008particle}          &   PSO & Multi-population & Adaptive & Adaptive & By index & Homogeneous\\
\rowcolor{lightgray}AMSO &\cite{li2014adaptive}&PSO & Multi-population & Adaptive & Adaptive & By position & Heterogeneous\\   
CDE& \cite{du2012using} & DE/best/2/bin & Multi-population & Fixed & Fixed & By index & Homogeneous\\
\rowcolor{lightgray}CESO& \cite{lung2007collaborative}& DE/rand/1/exp and PSO & Bi-population & Fixed & Fixed & N/A  & Heterogeneous\\
CPSO &\cite{yang2010clustering}&PSO& Multi-population & Adaptive & Fixed & By position & Heterogeneous\\
\rowcolor{lightgray}CPSOR &\cite{li2012general}&PSO & Multi-population & Adaptive & Fixed & By position & Heterogeneous\\   
DSPSO &\cite{parrot2006locating} &PSO & Multi-population & Adaptive & Fixed & By position and fitness & Heterogeneous\\ 
\rowcolor{lightgray}DynDE& \cite{mendes2006DynDE}&DE/best/2/bin & Multi-population & Fixed & Fixed & By index & Homogeneous\\      
DynPopDE& \cite{plessis2013differential}&DE/best/2/bin& Multi-population & Adaptive & Adaptive & By index & Homogeneous\\           
\rowcolor{lightgray}FTMPSO& \cite{yazdani2013novel} &  PSO  & Multi-population & Adaptive & Adaptive & By index & Heterogeneous\\
HmSO &\cite{kamosi2010hibernating}&PSO & Multi-population & Fixed & Fixed & By index & Homogeneous\\   
\rowcolor{lightgray}IDSPSO &\cite{blackwell2008particle} &PSO & Multi-population & Adaptive & Fixed & By position and fitness & Heterogeneous\\ 
ImQSO       & \cite{kordestani2019note}   &   PSO& Multi-population & Fixed & Fixed & By index & Homogeneous\\   
\rowcolor{lightgray}mCMA-ES    & \cite{yazdani2019scaling}                   &   CMA-ES& Multi-population & Adaptive & Adaptive & By index & Homogeneous\\
mDE       & \cite{yazdani2019scaling}                  &   DE/best/2/bin& Multi-population & Adaptive & Adaptive & By index & Homogeneous\\      
\rowcolor{lightgray}mjDE       & \cite{yazdani2019scaling}                   &   jDE& Multi-population & Adaptive & Adaptive & By index & Homogeneous\\
mPSO   & \cite{yazdani2019scaling} &   PSO & Multi-population & Adaptive & Adaptive & By index & Homogeneous\\
\rowcolor{lightgray}mQSO       & \cite{blackwell2006multiswarm}   &   PSO & Multi-population & Fixed & Fixed & By index & Homogeneous\\        
psfNBC &\cite{luo2018identifying}&PSO& Multi-population & Adaptive & Fixed & By position and fitness & Homogeneous\\  
\rowcolor{lightgray}RPSO       & \cite{hu2002adiptive}                         &   PSO  & Single-population & Fixed & Fixed & N/A & N/A\\
SPSO$_\mathrm{AD+AP}$ &\cite{yazdani2023species}&PSO& Multi-population & Adaptive & Adaptive & By position and fitness & Homogeneous\\  

\rowcolor{lightgray}TMIPSO &\cite{wang2007triggered}&PSO& Bi-population & Fixed & Fixed & N/A & Heterogeneous \\   
  \bottomrule
  \end{tabular}
}\end{table}

To ensure fair comparisons in experiments, we have standardized certain aspects of the optimization components used in the EDOAs, meaning that some EDOAs in EDOLAB may differ slightly from their original versions. 
For all EDOAs that utilize PSO as their optimization component, we employ PSO with a constriction factor~\cite{eberhart2001comparing}. 
Additionally, DE/rand/2/bin~\cite{mendes2006DynDE} is used for most EDOAs that incorporate DE. 
In CESO~\cite{lung2007collaborative}, crowding DE (DE/rand/1/exp)~\cite{thomsen2004multimodal} is used to maintain global diversity; thus, we have retained this DE version. 
Similarly, in mjDE, the jDE~\cite{brest2006selfadaptive}, a well-known self-adaptive version of DE, is employed. Furthermore, to handle the box constraints~\cite{mezura2011constraint} (i.e., keeping the individuals/candidate solutions within search space boundaries), we apply the absorb boundary handling technique~\cite{helwig2012experimental,gandomi2012evolutionary} across all EDOAs.

We also assume that all EDOAs in EDOLAB are informed about environmental changes; therefore, change detection components are not included. 
As described in~\cite{branke2003designing}, environmental changes in many real-world DOPs are often \emph{visible}, with optimization algorithms being informed of these changes through other system components, such as agents, sensors, or the arrival of new entities (for example, new orders)~\cite{yazdani2021DOPsurveyPartA}. 
In such scenarios, the algorithms do not require a change detection component.

In the original versions of some EDOAs, internal parameters of benchmark generators, such as shift severity, were utilized by the algorithms. 
However, for fair and unbiased comparisons, problem instances must be treated as black boxes, and using such internal knowledge can disadvantage other EDOAs that do not have access to it. 
In EDOLAB, we employ the shift severity estimation method from~\cite{yazdani2018robust} for all EDOAs that require knowledge about shift severity. 
Furthermore, in EDOAs and components that originally required knowledge of the number of promising regions, we use the number of sub-populations instead~\cite{blackwell2008particle}.

\color{black}

To provide an initial understanding of the performance of the EDOAs included in EDOLAB, we have conducted experiments on various scenarios generated by the platform's four benchmark generators, specifically focusing on scenarios with different numbers of promising regions. 
The results of all 25 EDOAs across these scenarios are presented in the appendix. 
These results are intended to help users better understand the platform's capabilities and guide them in selecting appropriate algorithms and benchmark scenarios for their own experiments.

\section{Overview of Structure and Architecture of EDOLAB}
\label{sec:Overview}

EDOLAB is an open-source MATLAB platform.
It offers a modular architecture that provides users with both flexibility and ease of use. 
The platform is equipped with two main modules: \textit{Experimentation} and \textit{Education}, each designed to meet the varying needs of researchers working on DOPs. 
EDOLAB also emphasizes extendibility, which allows users to incorporate new algorithms, benchmark generators, and performance indicators into its framework with minimal effort.

The \textit{Experimentation} module is a key component for conducting experiments and comparing the performance of different EDOAs across a wide range of problem instances. 
Users can configure a variety of parameters for both the algorithms and benchmark generators, making EDOLAB a powerful tool for benchmarking and algorithm evaluation. 
The experimentation module allows researchers to select from 25 pre-included EDOAs and four highly configurable benchmark generators. 
The module is available in both graphical (GUI) and non-graphical (script-based) modes, giving users the flexibility to choose between a more visual setup or advanced control via direct code manipulation.

The \textit{Education} module serves as a valuable tool for understanding the behaviors of EDOAs over time. 
It is designed to visualize the evolution of solutions and environmental changes, making it easier for researchers, particularly less experienced ones, to observe how dynamic optimization algorithms track the moving optimum across successive environments. 
This module provides 2-dimensional contour plots, displaying how individuals move in the search space, how changes in the environment affect the search process, and how EDOAs respond to these changes.

One of the most significant advantages of EDOLAB is its extendibility. 
The platform's architecture is designed to be easily expandable, allowing users to add their own EDOAs, benchmark generators, and performance indicators. 
Researchers can create new algorithms by following a few straightforward steps, such as adding a new sub-folder for the algorithm and defining its main function, ensuring it integrates smoothly with the rest of the platform. 
Similarly, adding new benchmark generators or performance indicators requires minimal changes to the source code, making EDOLAB a flexible tool for experimenting with custom-designed components or integrating state-of-the-art algorithms.

For users seeking to learn more about EDOLAB, the platform is accompanied by a comprehensive user manual that provides in-depth information on its architecture, including a sequence diagram illustrating how EDOLAB operates. 
The manual contains visual aids to help users navigate through various modules and the GUI, as well as step-by-step instructions for setting up experiments, configuring the platform, and extending it with new features such as new EDOAs, benchmark generators, and performance indicators. 
It covers both GUI and non-GUI modes, enabling users to efficiently run experiments, generate outputs, and analyze results based on their preferences. 
Practical examples are included to help users quickly familiarize themselves with the platform's capabilities.
The manual also provides instructions for users who prefer an open-source alternative or do not have access to a MATLAB license, outlining the necessary modifications for using EDOLAB in Octave while preserving the platform's main functionalities.

\color{black}

\section{Conclusion}
\label{sec:conclusion}

Evolutionary dynamic optimization algorithms (EDOAs) and dynamic benchmark generators are typically complex, making their re-implementation both challenging and prone to errors.
Over the past two decades, the absence of publicly available source codes for many EDOAs and benchmark generators has posed a significant challenge for researchers attempting to reproduce results for experimentation and comparison. 
To address this issue, we introduced EDOLAB, an open-source MATLAB platform designed for evolutionary dynamic optimization.

EDOLAB aims to assist researchers, especially those with less experience, in two primary ways: first, by helping them understand how EDOAs function and how dynamic benchmark instances exhibit various morphological and dynamical characteristics; and second, by facilitating the experimentation and comparison of EDOAs with other algorithms. 
These objectives are met through EDOLAB's two main modules--the \textit{Education} module and the \textit{Experimentation} module.

The initial release of EDOLAB includes 25 EDOAs, four highly configurable and parametric benchmark generators, and two commonly used performance indicators. 
In this paper, we described the technical aspects of EDOLAB, including its architecture, the process of running it with and without the GUI, the features of both modules, and the methods for extending the platform by adding new EDOAs, benchmark generators, and performance indicators.

As future work, expanding EDOLAB's library with more state-of-the-art EDOAs will be an important step in enriching the platform's capabilities. 
Additionally, extending EDOLAB to cover other key sub-fields of dynamic optimization---such as dynamic multi-objective optimization~\cite{jiang2022evolutionary,raquel2013dynamic}, large-scale dynamic optimization~\cite{yazdani2019scaling,bai2022evolutionary}, dynamic multimodal optimization~\cite{lin2022popdmmo,luo2019clonal}, dynamic constrained optimization~\cite{nguyen2012continuous,bu2016continuous}, and robust optimization over time~\cite{fu2015robust,yu2010robust,jin2013framework}---will be essential for future development.
{\color{black}
Another future work is the introduction of a centralized options structure for managing algorithm parameters and components, which would further improve the platform's user-friendliness for those who want to study the impact of different parameter settings and components on the performance of the algorithms.
Finally, re-implementing EDOLAB in Python is another future direction. 
As a free and widely used language, Python would make the platform more accessible and open it up to a broader audience.}

\color{black}

\part*{\LARGE Appendix 1: Experimental Results}

In this appendix, we present the experimental results for the Evolutionary Dynamic Optimization Algorithms (EDOAs) listed in Table~\ref{tab:algorithms}. 
These algorithms were evaluated using the four benchmark generators described in Section~\ref{sec:sec:benchmarks}.
For each benchmark generator, we ran the algorithms on dynamic optimization problem instances with varying numbers of promising regions ($\mathrm{P} \in \{5, 10, 25, 50, 100\}$). 
The dimension was set to 5, the change frequency was set to 5000, and the shift severity was set to 1 for all experiments. 
Each experiment was repeated for 31 independent runs.
The reported metrics include the following:

\begin{itemize}
    \item {Offline Error:} The average ($E_\mathrm{O}^\mathrm{avg}$), median ($E_\mathrm{O}^\mathrm{med}$), and standard deviation ($E_\mathrm{O}^\mathrm{std}$) of the offline error metric over 31 runs.
    \item {Average error before environmental changes:} The average ($E_\mathrm{BBC}^\mathrm{avg}$), median ($E_\mathrm{BBC}^\mathrm{med}$), and standard deviation ($E_\mathrm{BBC}^\mathrm{std}$) of the average error before environmental changes metric over 31 runs.
\end{itemize}

These experiments provide insights into the behavior of different EDOAs across a range of scenarios. Specifically, they help researchers understand how each algorithm performs in environments with different morphological characteristics (e.g., the number of promising regions). By evaluating the results of these experiments, users can identify the strengths and weaknesses of each algorithm in solving dynamic optimization problem instances. 
This can guide users in selecting a subset of algorithms and problem scenarios most appropriate for their own experiments.

Moreover, these results demonstrate the utility of EDOLAB as a platform for systematically comparing algorithms across various benchmarks and scenarios, promoting the reproducibility of experiments and facilitating further research in evolutionary dynamic optimization.
The detailed results of these experiments are provided in Tables~\ref{tab:MPBresults},~\ref{tab:GDBGresults},~\ref{tab:FPsresults}, and~\ref{tab:GMPBresults}.

\newpage
\newgeometry{ 
  top=15mm,
  bottom=15mm,
  left=25mm,
  right=25mm
}
\renewcommand{\thetable}{A1-\arabic{table}} 
\setcounter{table}{0}
\renewcommand{\arraystretch}{1.7} 
\setlength{\tabcolsep}{3.3pt}

\begin{table}[!t]
\color{black}
\centering
  \caption{\color{black}
Performance statistics of algorithms on the MPB benchmark, evaluated across different instances with 5, 10, 25, 50, and 100 promising regions. The table shows the average, median, and standard deviation of the offline error and the average best error before environmental changes, based on 31 runs for each scenario. }
  \label{tab:MPBresults}
\Rotatebox{90}{
\tiny
    \begin{tabular}{c|c|ccccccccccccccccccccccccc}
      \hline
      \hline
    P     & I & ACFPSO & AMPDE & AMPPSO & AMSO & AmQSO & CDE & CESO & CPSO & CPSOR & DSPSO & DynDE & DynPopDE & FTMPSO & HmSO & IDSPSO & ImQSO & RPSO & SPSO$_\mathrm{AD+AP}$ & TMIPSO & mCMAES & mDE & mPSO & mQSO & mjDE & psfNBC \\
      \hline
      \multirow{6}[0]{*}{5} & $E_\mathrm{O}^\mathrm{avg}$ & 0.777348 & 1.798459 & 1.311168 & 1.504578 & 0.932503 & 1.810491 & 11.416614 & 3.973526 & 2.727747 & 4.139301 & 1.89893 & 2.104509 & 0.826357 & 2.011164 & 1.519715 & 2.086776 & 12.622708 & 0.875273 & 6.123402 & 1.470653 & 1.715075 & 0.979271 & 2.078159 & 3.212291 & 2.372827 \\
      & $E_\mathrm{O}^\mathrm{med}$ & 0.694257 & 1.488037 & 1.157462 & 1.452006 & 0.808436 & 1.751537 & 11.263262 & 3.938418 & 2.593088 & 3.685273 & 1.674708 & 1.874268 & 0.598182 & 1.654453 & 1.368183 & 2.037181 & 13.492024 & 0.607328 & 5.701739 & 1.088505 & 1.596607 & 0.719402 & 2.036125 & 2.946061 & 2.190052 \\
      & $E_\mathrm{O}^\mathrm{std}$ & 0.061366 & 0.143442 & 0.099037 & 0.067221 & 0.070588 & 0.067013 & 0.866451 & 0.131772 & 0.132572 & 0.290923 & 0.132033 & 0.165589 & 0.095209 & 0.171348 & 0.102858 & 0.08031 & 0.571869 & 0.097985 & 0.404043 & 0.20678 & 0.109162 & 0.088685 & 0.082613 & 0.243189 & 0.113943 \\
      & $E_\mathrm{BBC}^\mathrm{avg}$ & 0.345732 & 0.629227 & 0.518586 & 0.245401 & 0.349507 & 0.473962 & 11.032342 & 0.961007 & 1.144387 & 2.42981 & 0.609955 & 1.449204 & 0.481555 & 1.126233 & 0.522444 & 0.698034 & 12.197224 & 0.480489 & 4.880301 & 1.019976 & 1.0975 & 0.410154 & 0.651008 & 2.024649 & 0.780893 \\
      & $E_\mathrm{BBC}^\mathrm{med}$ & 0.27617 & 0.274611 & 0.327907 & 0.092408 & 0.190467 & 0.361774 & 10.803412 & 0.822624 & 0.911332 & 2.070168 & 0.387946 & 1.340637 & 0.278289 & 0.910448 & 0.33198 & 0.553474 & 12.842863 & 0.245308 & 4.53065 & 0.622206 & 0.872202 & 0.170439 & 0.563675 & 1.728255 & 0.527108 \\
      & $E_\mathrm{BBC}^\mathrm{std}$ & 0.061472 & 0.153366 & 0.102283 & 0.057465 & 0.073666 & 0.057746 & 0.882662 & 0.091301 & 0.126442 & 0.299578 & 0.127189 & 0.165365 & 0.098004 & 0.175581 & 0.098482 & 0.06042 & 0.576359 & 0.097123 & 0.395309 & 0.208961 & 0.104446 & 0.084927 & 0.055209 & 0.251305 & 0.110281 \\
        \hline
      \multirow{6}[0]{*}{10} & $E_\mathrm{O}^\mathrm{avg}$ & 0.978274 & 3.022642 & 1.74132 & 1.737637 & 1.306027 & 1.633979 & 15.393587 & 3.69739 & 2.465276 & 6.012556 & 2.132655 & 2.368593 & 0.952999 & 2.495978 & 1.631366 & 2.226377 & 14.594914 & 1.029411 & 5.570382 & 2.252811 & 1.907507 & 1.366195 & 2.093027 & 3.702012 & 2.548607 \\
      & $E_\mathrm{O}^\mathrm{med}$ & 0.982087 & 2.091393 & 1.587775 & 1.629687 & 1.283199 & 1.547832 & 16.18503 & 3.609438 & 2.47332 & 5.671263 & 1.980958 & 2.143492 & 0.951066 & 2.419725 & 1.535944 & 2.201705 & 14.110859 & 0.953314 & 5.580694 & 1.942669 & 1.76448 & 1.368228 & 2.043826 & 3.39984 & 2.545304 \\
      & $E_\mathrm{O}^\mathrm{std}$ & 0.055464 & 0.374898 & 0.098551 & 0.097335 & 0.069534 & 0.097432 & 0.828628 & 0.105619 & 0.073222 & 0.420991 & 0.108719 & 0.13136 & 0.066427 & 0.155569 & 0.089562 & 0.080169 & 0.595138 & 0.064905 & 0.361887 & 0.176219 & 0.077217 & 0.065446 & 0.079388 & 0.219103 & 0.082865 \\
      & $E_\mathrm{BBC}^\mathrm{avg}$ & 0.477906 & 1.975894 & 0.847459 & 0.667048 & 0.547907 & 0.707384 & 15.069901 & 1.1887 & 1.109086 & 4.603455 & 1.078775 & 1.692907 & 0.575044 & 1.626063 & 0.787805 & 1.068393 & 14.17829 & 0.560218 & 4.54752 & 1.737992 & 1.243648 & 0.611056 & 0.94157 & 2.567563 & 1.076765 \\
      & $E_\mathrm{BBC}^\mathrm{med}$ & 0.440073 & 1.037923 & 0.66203 & 0.531494 & 0.484871 & 0.545925 & 15.806547 & 1.129651 & 1.128207 & 4.106716 & 0.922833 & 1.470477 & 0.537596 & 1.596274 & 0.67169 & 0.92682 & 13.884415 & 0.449176 & 4.432044 & 1.602919 & 1.14347 & 0.534931 & 0.852187 & 2.142208 & 0.878872 \\
      & $E_\mathrm{BBC}^\mathrm{std}$ & 0.05472 & 0.394761 & 0.099394 & 0.091778 & 0.072325 & 0.090153 & 0.837447 & 0.076387 & 0.074264 & 0.412508 & 0.100095 & 0.134543 & 0.063744 & 0.15151 & 0.088126 & 0.079108 & 0.596364 & 0.06936 & 0.349418 & 0.179612 & 0.07301 & 0.071644 & 0.074238 & 0.217511 & 0.079142 \\
        \hline
      \multirow{6}[0]{*}{25} & $E_\mathrm{O}^\mathrm{avg}$ & 1.40184 & 2.862828 & 2.164541 & 2.11366 & 1.894259 & 2.778031 & 14.113405 & 3.553832 & 2.687233 & 8.022354 & 3.175468 & 2.604148 & 1.243923 & 2.761284 & 2.407354 & 2.753639 & 12.894239 & 1.394924 & 4.846211 & 2.270445 & 2.298269 & 1.879574 & 2.713434 & 4.040259 & 3.249288 \\
      & $E_\mathrm{O}^\mathrm{med}$ & 1.458245 & 2.742258 & 2.11845 & 2.067586 & 1.910399 & 2.595453 & 13.089373 & 3.497717 & 2.714499 & 7.70315 & 3.177264 & 2.575702 & 1.215825 & 2.692014 & 2.318465 & 2.664807 & 12.501926 & 1.395421 & 5.072457 & 2.164384 & 2.267302 & 1.844697 & 2.707197 & 3.960015 & 3.291138 \\
      & $E_\mathrm{O}^\mathrm{std}$ & 0.052907 & 0.133246 & 0.077593 & 0.079895 & 0.066588 & 0.092693 & 0.88459 & 0.103785 & 0.09102 & 0.328259 & 0.114467 & 0.099543 & 0.05305 & 0.120617 & 0.103694 & 0.089957 & 0.491667 & 0.051387 & 0.159201 & 0.096848 & 0.069014 & 0.066189 & 0.088297 & 0.141401 & 0.090903 \\
      & $E_\mathrm{BBC}^\mathrm{avg}$ & 0.786716 & 1.745331 & 1.208135 & 1.120404 & 0.945223 & 2.008437 & 13.742083 & 1.62417 & 1.654461 & 6.755098 & 2.232085 & 1.78464 & 0.727216 & 1.939386 & 1.656197 & 1.779017 & 12.408415 & 0.793308 & 3.876848 & 1.495375 & 1.489139 & 0.931407 & 1.733277 & 2.840772 & 1.905433 \\
      & $E_\mathrm{BBC}^\mathrm{med}$ & 0.762204 & 1.514836 & 1.076578 & 0.994771 & 0.936937 & 1.799119 & 12.827639 & 1.567646 & 1.581997 & 6.762802 & 2.19638 & 1.695833 & 0.719219 & 1.871861 & 1.475476 & 1.614347 & 11.983377 & 0.748444 & 3.971466 & 1.39214 & 1.45602 & 0.844675 & 1.734174 & 2.75024 & 1.837359 \\
      & $E_\mathrm{BBC}^\mathrm{std}$ & 0.051724 & 0.139354 & 0.080948 & 0.090461 & 0.061599 & 0.087613 & 0.893541 & 0.081256 & 0.082889 & 0.337188 & 0.107412 & 0.093458 & 0.050835 & 0.117821 & 0.099351 & 0.07998 & 0.494938 & 0.043868 & 0.151502 & 0.096045 & 0.064784 & 0.063143 & 0.07852 & 0.129963 & 0.084014 \\
        \hline
      \multirow{6}[0]{*}{50} & $E_\mathrm{O}^\mathrm{avg}$ & 1.555488 & 3.072901 & 2.326177 & 2.141367 & 2.083622 & 3.448897 & 14.037977 & 3.102864 & 2.545048 & 11.415602 & 3.682847 & 2.492632 & 1.357667 & 2.657126 & 2.523762 & 2.678016 & 11.676747 & 1.55941 & 4.111637 & 2.217194 & 2.396736 & 2.09536 & 2.630544 & 3.929258 & 3.173255 \\
      & $E_\mathrm{O}^\mathrm{med}$ & 1.518225 & 2.629936 & 2.312205 & 2.115007 & 2.080679 & 3.243536 & 15.120511 & 3.096661 & 2.555485 & 11.196656 & 3.528229 & 2.463091 & 1.35403 & 2.514612 & 2.551899 & 2.504047 & 11.640249 & 1.510355 & 3.946521 & 2.171979 & 2.346018 & 2.096302 & 2.619584 & 3.901084 & 3.134736 \\
      & $E_\mathrm{O}^\mathrm{std}$ & 0.042047 & 0.197293 & 0.050489 & 0.059977 & 0.046373 & 0.109435 & 0.950172 & 0.064045 & 0.067814 & 0.566783 & 0.122421 & 0.068781 & 0.035127 & 0.084316 & 0.084465 & 0.077133 & 0.362203 & 0.042002 & 0.112718 & 0.063063 & 0.051351 & 0.040802 & 0.063867 & 0.106182 & 0.054952 \\
      & $E_\mathrm{BBC}^\mathrm{avg}$ & 0.906816 & 2.003474 & 1.365798 & 1.095423 & 1.082307 & 2.732044 & 13.61987 & 1.532324 & 1.676965 & 10.358878 & 2.766268 & 1.616883 & 0.707578 & 1.867709 & 1.845785 & 1.851282 & 11.178544 & 0.928369 & 3.124402 & 1.331696 & 1.515693 & 1.102883 & 1.801942 & 2.754195 & 1.91865 \\
      & $E_\mathrm{BBC}^\mathrm{med}$ & 0.866903 & 1.466183 & 1.320435 & 1.072305 & 1.071771 & 2.551033 & 14.803681 & 1.55632 & 1.714816 & 10.153125 & 2.629581 & 1.57975 & 0.692781 & 1.751947 & 1.878 & 1.713735 & 10.931583 & 0.910853 & 3.014411 & 1.295293 & 1.49571 & 1.08248 & 1.777993 & 2.662333 & 1.885039 \\
      & $E_\mathrm{BBC}^\mathrm{std}$ & 0.03965 & 0.19781 & 0.042055 & 0.060969 & 0.037326 & 0.109571 & 0.974862 & 0.047372 & 0.064346 & 0.598713 & 0.122809 & 0.061964 & 0.03191 & 0.074032 & 0.077467 & 0.069741 & 0.359327 & 0.039838 & 0.109733 & 0.061934 & 0.041761 & 0.036306 & 0.058268 & 0.095062 & 0.045143 \\
        \hline
      \multirow{6}[0]{*}{100} & $E_\mathrm{O}^\mathrm{avg}$ & 1.534706 & 3.299049 & 2.411868 & 2.284824 & 2.166232 & 3.732919 & 12.395075 & 2.676737 & 2.636708 & 11.235644 & 3.868488 & 2.511524 & 1.408738 & 2.541729 & 2.432752 & 2.578544 & 10.722634 & 1.53912 & 3.69927 & 2.162951 & 2.415937 & 2.216141 & 2.502395 & 3.894097 & 3.140181 \\
      & $E_\mathrm{O}^\mathrm{med}$ & 1.554145 & 3.030366 & 2.28765 & 2.214088 & 2.15468 & 3.448238 & 10.793637 & 2.617693 & 2.685351 & 11.313579 & 3.774169 & 2.486827 & 1.411355 & 2.469591 & 2.457829 & 2.582558 & 10.695901 & 1.53021 & 3.771916 & 2.155744 & 2.386915 & 2.203743 & 2.482888 & 3.764688 & 3.157307 \\
      & $E_\mathrm{O}^\mathrm{std}$ & 0.027965 & 0.215829 & 0.097039 & 0.065905 & 0.031047 & 0.143154 & 0.910414 & 0.041415 & 0.05315 & 0.373277 & 0.111501 & 0.04895 & 0.022003 & 0.073773 & 0.051953 & 0.044906 & 0.296841 & 0.034414 & 0.063397 & 0.039522 & 0.031599 & 0.030849 & 0.047 & 0.105733 & 0.049152 \\
      & $E_\mathrm{BBC}^\mathrm{avg}$ & 0.930852 & 2.277594 & 1.532352 & 1.274552 & 1.222699 & 3.06202 & 11.931452 & 1.375511 & 1.794177 & 10.258792 & 3.010032 & 1.598305 & 0.717216 & 1.775438 & 1.802216 & 1.808791 & 10.226951 & 0.884653 & 2.701052 & 1.257693 & 1.481136 & 1.254565 & 1.750419 & 2.798599 & 1.95189 \\
      & $E_\mathrm{BBC}^\mathrm{med}$ & 0.920692 & 1.9566 & 1.390685 & 1.169888 & 1.233099 & 2.846082 & 10.300354 & 1.347878 & 1.834432 & 10.4244 & 3.039576 & 1.562257 & 0.728863 & 1.715788 & 1.811971 & 1.799051 & 10.26644 & 0.85991 & 2.679231 & 1.237988 & 1.456209 & 1.24166 & 1.773103 & 2.703047 & 1.950612 \\
      & $E_\mathrm{BBC}^\mathrm{std}$ & 0.025962 & 0.213419 & 0.099598 & 0.068925 & 0.022528 & 0.135175 & 0.921123 & 0.030424 & 0.04797 & 0.394522 & 0.108154 & 0.043874 & 0.015971 & 0.064884 & 0.046016 & 0.040902 & 0.295777 & 0.032495 & 0.055494 & 0.030537 & 0.022888 & 0.021262 & 0.043471 & 0.096135 & 0.042711 \\
      \hline
      \hline
    \end{tabular}%
  }
\end{table}

\begin{table}[!t]
\color{black}
\centering
  \caption{\color{black}
Performance statistics of algorithms on the GDBG benchmark, evaluated across different instances with 5, 10, 25, 50, and 100 promising regions. The table shows the average, median, and standard deviation of the offline error and the average best error before environmental changes, based on 31 runs for each scenario. }
  \label{tab:GDBGresults}
\Rotatebox{90}{
\tiny
    \begin{tabular}{c|c|ccccccccccccccccccccccccc}
      \hline
      \hline
    P     & I & ACFPSO & AMPDE & AMPPSO & AMSO & AmQSO & CDE & CESO & CPSO & CPSOR & DSPSO & DynDE & DynPopDE & FTMPSO & HmSO & IDSPSO & ImQSO & RPSO & SPSO$_\mathrm{AD+AP}$ & TMIPSO & mCMAES & mDE & mPSO & mQSO & mjDE & psfNBC \\
      \hline
      \multirow{6}[0]{*}{5} & $E_\mathrm{O}^\mathrm{avg}$ & 5.441255 & 8.017961 & 5.833218 & 6.887193 & 6.619299 & 9.479417 & 30.044232 & 16.456456 & 8.635112 & 44.235487 & 10.231639 & 11.203819 & 6.366048 & 8.508585 & 6.887206 & 14.2105 & 35.424692 & 6.818792 & 22.583672 & 4.11482 & 8.237566 & 6.471514 & 14.029444 & 23.405479 & 15.421975 \\
       & $E_\mathrm{O}^\mathrm{med}$ & 5.343704 & 7.702214 & 5.907756 & 7.045997 & 6.514563 & 9.378745 & 30.393218 & 16.086524 & 8.617817 & 36.305982 & 8.863541 & 11.024621 & 6.338674 & 8.283925 & 6.869292 & 14.148916 & 34.819608 & 7.00857 & 21.854325 & 3.957523 & 7.995599 & 6.406547 & 14.116494 & 24.206748 & 15.550235 \\
       & $E_\mathrm{O}^\mathrm{std}$ & 0.229287 & 0.41009 & 0.200863 & 0.170088 & 0.269274 & 0.436611 & 1.192216 & 0.379064 & 0.204632 & 5.296181 & 0.825437 & 0.492959 & 0.232273 & 0.303674 & 0.291263 & 0.452839 & 1.712309 & 0.229768 & 0.595715 & 0.187897 & 0.268015 & 0.250814 & 0.428076 & 1.003844 & 0.392395 \\
       & $E_\mathrm{BBC}^\mathrm{avg}$ & 2.186502 & 3.042007 & 1.903825 & 2.01555 & 2.309048 & 4.47894 & 24.609324 & 6.104166 & 2.929096 & 37.085579 & 4.343283 & 6.258177 & 4.046278 & 3.149109 & 2.751761 & 7.845877 & 29.486336 & 4.983458 & 15.469208 & 1.607193 & 5.084089 & 2.283914 & 7.577798 & 12.849192 & 7.330249 \\
       & $E_\mathrm{BBC}^\mathrm{med}$ & 2.116574 & 2.559656 & 1.861152 & 1.911041 & 2.168897 & 3.7742 & 24.697219 & 5.963168 & 2.74998 & 30.297994 & 2.860607 & 6.46494 & 4.045623 & 3.16636 & 2.6182 & 7.967276 & 29.942635 & 4.982463 & 15.021497 & 1.436374 & 5.124806 & 2.304936 & 7.31725 & 13.172602 & 7.485414 \\
       & $E_\mathrm{BBC}^\mathrm{std}$ & 0.157529 & 0.299255 & 0.105624 & 0.099437 & 0.156728 & 0.355074 & 1.132243 & 0.201719 & 0.156839 & 5.427772 & 0.708383 & 0.26693 & 0.170714 & 0.207028 & 0.164695 & 0.35967 & 1.68214 & 0.181746 & 0.472777 & 0.155753 & 0.201855 & 0.129626 & 0.323752 & 0.62152 & 0.25613 \\
         \hline
       \multirow{6}[0]{*}{10} & $E_\mathrm{O}^\mathrm{avg}$ & 3.689678 & 5.693045 & 4.783161 & 3.801665 & 4.974793 & 5.041022 & 32.257591 & 11.307115 & 6.06003 & 41.537634 & 5.523024 & 6.601605 & 4.311363 & 7.121573 & 4.671339 & 9.490018 & 34.405741 & 5.78535 & 14.684719 & 3.454275 & 6.371388 & 5.042612 & 9.159472 & 19.611845 & 11.34719 \\
       & $E_\mathrm{O}^\mathrm{med}$ & 3.571844 & 5.011155 & 4.641235 & 3.624055 & 4.727248 & 4.936804 & 32.122191 & 10.917321 & 5.87279 & 40.096855 & 5.773222 & 6.328826 & 4.15105 & 6.874366 & 4.549135 & 9.137705 & 34.516803 & 5.619851 & 14.449479 & 3.412162 & 6.191907 & 4.772479 & 8.996904 & 18.934172 & 11.363571 \\
       & $E_\mathrm{O}^\mathrm{std}$ & 0.11578 & 0.280221 & 0.136729 & 0.147222 & 0.147775 & 0.166032 & 1.289987 & 0.223653 & 0.164089 & 2.201512 & 0.164189 & 0.239515 & 0.147031 & 0.236095 & 0.134286 & 0.27954 & 1.019619 & 0.190406 & 0.420554 & 0.107259 & 0.188696 & 0.193617 & 0.260532 & 0.421274 & 0.219152 \\
       & $E_\mathrm{BBC}^\mathrm{avg}$ & 1.486951 & 2.129481 & 1.910028 & 1.110827 & 1.801082 & 2.573263 & 26.82304 & 5.329389 & 2.510263 & 36.520456 & 2.110346 & 3.658866 & 2.664898 & 3.051432 & 2.003418 & 5.516311 & 29.150536 & 3.93318 & 11.108939 & 1.048781 & 3.822409 & 1.805187 & 5.401686 & 11.902205 & 5.183932 \\
       & $E_\mathrm{BBC}^\mathrm{med}$ & 1.341462 & 1.735453 & 1.862422 & 1.020765 & 1.818974 & 2.517234 & 26.639705 & 5.223897 & 2.463305 & 32.765281 & 2.01394 & 3.533038 & 2.704183 & 2.81251 & 2.079047 & 5.258178 & 29.500778 & 3.785752 & 11.097514 & 1.087187 & 3.752884 & 1.704306 & 5.315113 & 11.714698 & 4.947171 \\
       & $E_\mathrm{BBC}^\mathrm{std}$ & 0.082716 & 0.217123 & 0.088219 & 0.099049 & 0.086945 & 0.127734 & 1.184131 & 0.152254 & 0.109278 & 2.24108 & 0.115885 & 0.1425 & 0.132553 & 0.18848 & 0.094123 & 0.221873 & 0.994311 & 0.16206 & 0.338456 & 0.071364 & 0.141671 & 0.108237 & 0.205235 & 0.295534 & 0.132288 \\
         \hline
       \multirow{6}[0]{*}{25} & $E_\mathrm{O}^\mathrm{avg}$ & 3.056765 & 5.71685 & 4.264203 & 3.538291 & 4.222549 & 6.443949 & 31.545261 & 7.523695 & 4.575121 & 38.62054 & 6.459048 & 4.792524 & 3.058162 & 6.154769 & 5.503661 & 8.8947 & 27.420145 & 5.881313 & 9.713702 & 3.556781 & 4.869432 & 4.212174 & 8.684062 & 15.1098 & 9.692322 \\
       & $E_\mathrm{O}^\mathrm{med}$ & 3.000895 & 5.350566 & 4.21702 & 3.431784 & 4.212002 & 6.373392 & 31.987354 & 7.367728 & 4.539697 & 37.786455 & 6.618034 & 4.740715 & 3.114268 & 6.080649 & 5.442558 & 8.657924 & 27.076674 & 5.780156 & 9.38421 & 3.57307 & 4.841872 & 4.257635 & 8.613872 & 15.171528 & 9.648571 \\
       & $E_\mathrm{O}^\mathrm{std}$ & 0.078213 & 0.380388 & 0.108841 & 0.080704 & 0.089552 & 0.170782 & 0.982937 & 0.135489 & 0.075361 & 2.171943 & 0.15991 & 0.103288 & 0.070833 & 0.118043 & 0.148612 & 0.172396 & 0.649807 & 0.13409 & 0.199737 & 0.083529 & 0.108478 & 0.100214 & 0.155229 & 0.20489 & 0.14713 \\
       & $E_\mathrm{BBC}^\mathrm{avg}$ & 1.450757 & 2.742411 & 2.054131 & 1.330017 & 1.914284 & 4.214621 & 26.68924 & 4.028782 & 2.137426 & 36.30881 & 3.498775 & 2.497111 & 1.55253 & 2.734558 & 3.024127 & 5.015711 & 23.463757 & 3.338184 & 6.46622 & 1.417021 & 2.720965 & 1.934909 & 4.860683 & 8.233308 & 4.391465 \\
       & $E_\mathrm{BBC}^\mathrm{med}$ & 1.390647 & 2.350714 & 1.974769 & 1.25761 & 1.831826 & 4.120722 & 26.765425 & 3.932974 & 2.077743 & 35.563088 & 3.423461 & 2.523716 & 1.554674 & 2.710289 & 2.819171 & 4.897204 & 23.143016 & 3.340514 & 6.402011 & 1.381661 & 2.694613 & 1.862784 & 4.868392 & 8.13944 & 4.373194 \\
       & $E_\mathrm{BBC}^\mathrm{std}$ & 0.061747 & 0.28801 & 0.076644 & 0.060008 & 0.062886 & 0.150475 & 0.868682 & 0.09297 & 0.058734 & 2.2584 & 0.128813 & 0.068015 & 0.059353 & 0.085918 & 0.11491 & 0.124384 & 0.625652 & 0.095498 & 0.173659 & 0.059482 & 0.071608 & 0.064554 & 0.13879 & 0.184911 & 0.120611 \\
         \hline
       \multirow{6}[0]{*}{50} & $E_\mathrm{O}^\mathrm{avg}$ & 3.309678 & 5.078391 & 4.372141 & 3.998455 & 4.129254 & 6.704264 & 30.28925 & 6.348088 & 4.518101 & 34.197627 & 6.933424 & 4.831624 & 2.966697 & 5.828509 & 6.352245 & 8.777784 & 25.13275 & 6.983222 & 8.277563 & 3.896102 & 4.532362 & 4.175181 & 8.756632 & 15.040079 & 8.72562 \\
       & $E_\mathrm{O}^\mathrm{med}$ & 3.299192 & 4.892945 & 4.429169 & 3.937482 & 4.053415 & 6.561171 & 29.326477 & 6.412298 & 4.396162 & 32.969439 & 6.691207 & 4.822615 & 2.918082 & 5.821682 & 6.1694 & 9.013023 & 25.029251 & 6.867939 & 8.115786 & 3.902502 & 4.44646 & 4.014357 & 8.667046 & 15.106234 & 8.835534 \\
       & $E_\mathrm{O}^\mathrm{std}$ & 0.068231 & 0.235872 & 0.076961 & 0.081106 & 0.087877 & 0.15237 & 0.862098 & 0.089301 & 0.081334 & 0.984406 & 0.179213 & 0.095377 & 0.047447 & 0.087485 & 0.167233 & 0.165842 & 0.445527 & 0.154481 & 0.136871 & 0.076463 & 0.101751 & 0.085812 & 0.15163 & 0.246462 & 0.13205 \\
       & $E_\mathrm{BBC}^\mathrm{avg}$ & 1.747331 & 2.604884 & 2.430074 & 1.906701 & 2.366754 & 4.543898 & 25.371524 & 3.623898 & 2.247186 & 32.69011 & 4.039631 & 2.676762 & 1.436735 & 2.70956 & 3.82633 & 4.902513 & 21.780525 & 4.060527 & 4.929001 & 1.905144 & 2.791179 & 2.357508 & 4.843761 & 8.108582 & 3.866138 \\
       & $E_\mathrm{BBC}^\mathrm{med}$ & 1.745084 & 2.468364 & 2.494104 & 1.884296 & 2.373498 & 4.405635 & 24.654241 & 3.714406 & 2.208708 & 31.854945 & 3.845616 & 2.703163 & 1.434581 & 2.680821 & 3.725271 & 4.891337 & 21.824595 & 3.96407 & 4.887504 & 1.890188 & 2.753936 & 2.345567 & 4.760991 & 8.078439 & 3.860592 \\
       & $E_\mathrm{BBC}^\mathrm{std}$ & 0.056358 & 0.15429 & 0.055167 & 0.065753 & 0.061232 & 0.138648 & 0.83266 & 0.066483 & 0.061618 & 0.976072 & 0.14566 & 0.064307 & 0.035422 & 0.051583 & 0.138589 & 0.117796 & 0.397564 & 0.110568 & 0.097652 & 0.055765 & 0.081803 & 0.062491 & 0.103842 & 0.187676 & 0.086137 \\
         \hline
       \multirow{6}[0]{*}{100} & $E_\mathrm{O}^\mathrm{avg}$ & 3.541491 & 5.202949 & 4.45059 & 4.208868 & 4.135886 & 7.672019 & 31.636557 & 5.749378 & 4.767784 & 31.25489 & 7.673013 & 5.066752 & 2.776161 & 5.736414 & 6.655103 & 8.735994 & 23.022273 & 7.19989 & 7.538222 & 3.975857 & 4.440063 & 4.120124 & 8.727717 & 15.127364 & 8.119521 \\
       & $E_\mathrm{O}^\mathrm{med}$ & 3.542877 & 4.723545 & 4.457858 & 4.241848 & 4.131833 & 7.468316 & 32.59329 & 5.747591 & 4.77168 & 30.410681 & 7.605243 & 4.988063 & 2.75249 & 5.672916 & 6.589363 & 8.733019 & 23.511981 & 7.322935 & 7.604833 & 3.99331 & 4.423314 & 4.15759 & 8.609553 & 15.336196 & 8.133979 \\
       & $E_\mathrm{O}^\mathrm{std}$ & 0.056099 & 0.264072 & 0.092396 & 0.10931 & 0.084422 & 0.210747 & 1.021268 & 0.103414 & 0.110844 & 0.919051 & 0.214917 & 0.106291 & 0.054633 & 0.105277 & 0.182744 & 0.148096 & 0.419456 & 0.137098 & 0.109389 & 0.080419 & 0.101135 & 0.088745 & 0.142569 & 0.233741 & 0.112411 \\
       & $E_\mathrm{BBC}^\mathrm{avg}$ & 1.98839 & 2.809098 & 2.646184 & 2.154006 & 2.629073 & 5.449037 & 26.888451 & 3.268443 & 2.588684 & 30.261904 & 4.656028 & 2.894137 & 1.491046 & 2.715291 & 4.134004 & 4.811109 & 19.886786 & 4.065485 & 4.09558 & 2.175581 & 2.814295 & 2.620263 & 4.777538 & 8.624497 & 3.469673 \\
       & $E_\mathrm{BBC}^\mathrm{med}$ & 1.988326 & 2.534856 & 2.58282 & 2.199219 & 2.637848 & 5.351796 & 27.537175 & 3.371382 & 2.533248 & 29.356903 & 4.450739 & 2.874047 & 1.456821 & 2.700133 & 4.056508 & 4.76114 & 19.86918 & 4.095956 & 4.052667 & 2.135854 & 2.751824 & 2.606875 & 4.686322 & 8.940857 & 3.494733 \\
       & $E_\mathrm{BBC}^\mathrm{std}$ & 0.045259 & 0.143825 & 0.066817 & 0.063438 & 0.05943 & 0.192027 & 0.969769 & 0.081919 & 0.080244 & 0.91974 & 0.193881 & 0.071539 & 0.031792 & 0.073065 & 0.149483 & 0.107214 & 0.3997 & 0.102865 & 0.063832 & 0.05704 & 0.071101 & 0.062757 & 0.115394 & 0.181057 & 0.072195 \\
      \hline
      \hline
    \end{tabular}%
  }
\end{table}

\begin{table}[!t]
\color{black}
\centering
  \caption{\color{black}
Performance statistics of algorithms on the FPs benchmark, evaluated across different instances with 5, 10, 25, 50, and 100 promising regions. The table shows the average, median, and standard deviation of the offline error and the average best error before environmental changes, based on 31 runs for each scenario. }
  \label{tab:FPsresults}  
\Rotatebox{90}{
\tiny
    \begin{tabular}{c|c|ccccccccccccccccccccccccc}
      \hline
      \hline
    P     & I & ACFPSO & AMPDE & AMPPSO & AMSO & AmQSO & CDE & CESO & CPSO & CPSOR & DSPSO & DynDE & DynPopDE & FTMPSO & HmSO & IDSPSO & ImQSO & RPSO & SPSO$_\mathrm{AD+AP}$ & TMIPSO & mCMAES & mDE & mPSO & mQSO & mjDE & psfNBC \\
      \hline
    \multirow{6}[0]{*}{5} & $E_\mathrm{O}^\mathrm{avg}$ & 6.155632 & 6.279048 & 7.086594 & 6.657467 & 6.003462 & 8.46081 & 32.960646 & 10.426703 & 7.627212 & 50.705117 & 8.510369 & 8.564075 & 7.629901 & 9.062617 & 16.355382 & 9.441002 & 21.106498 & 6.93664 & 10.351155 & 6.499125 & 7.938344 & 5.800274 & 9.727037 & 12.930364 & 10.313816 \\
    & $E_\mathrm{O}^\mathrm{med}$ & 5.960786 & 6.111481 & 6.590816 & 6.819043 & 5.830184 & 8.095878 & 25.873262 & 10.450143 & 7.590523 & 51.700872 & 8.147885 & 8.639792 & 7.22812 & 9.045577 & 15.365461 & 9.086611 & 20.514469 & 7.067829 & 10.254116 & 6.108201 & 7.558439 & 5.509181 & 9.871456 & 13.161975 & 10.024273 \\
    & $E_\mathrm{O}^\mathrm{std}$ & 0.323806 & 0.440114 & 0.391222 & 0.265956 & 0.358772 & 0.380127 & 3.299897 & 0.458036 & 0.319954 & 1.124121 & 0.433514 & 0.391762 & 0.359664 & 0.434865 & 0.969362 & 0.430255 & 1.094528 & 0.428171 & 0.407345 & 0.291435 & 0.410267 & 0.338162 & 0.4149 & 0.609458 & 0.430105 \\
    & $E_\mathrm{BBC}^\mathrm{avg}$ & 4.240851 & 2.572718 & 3.948704 & 3.415669 & 3.091144 & 6.588876 & 30.211829 & 7.417805 & 4.445388 & 50.547029 & 5.691153 & 7.319727 & 6.490744 & 5.526428 & 13.276192 & 7.75914 & 19.915458 & 5.055294 & 8.661724 & 3.434417 & 6.345833 & 2.945494 & 8.075283 & 9.900912 & 8.162894 \\
    & $E_\mathrm{BBC}^\mathrm{med}$ & 3.797754 & 2.347846 & 3.289474 & 3.34834 & 2.910571 & 6.2878 & 21.908852 & 7.171273 & 4.365974 & 51.556446 & 5.287803 & 7.075094 & 6.071073 & 5.250041 & 12.463654 & 7.370922 & 19.519559 & 4.989741 & 8.665549 & 3.265445 & 6.106852 & 2.752238 & 7.994425 & 9.757909 & 8.18882 \\
    & $E_\mathrm{BBC}^\mathrm{std}$ & 0.227724 & 0.280068 & 0.349992 & 0.150012 & 0.23397 & 0.306011 & 3.411295 & 0.350302 & 0.187369 & 1.12448 & 0.372627 & 0.328189 & 0.311775 & 0.38213 & 0.963586 & 0.393069 & 1.098893 & 0.350367 & 0.388988 & 0.203047 & 0.369056 & 0.231261 & 0.389734 & 0.517655 & 0.384656 \\
      \hline
    \multirow{6}[0]{*}{10} & $E_\mathrm{O}^\mathrm{avg}$ & 5.133765 & 6.167375 & 6.285626 & 4.991945 & 5.571957 & 6.635679 & 37.964946 & 7.946682 & 6.164888 & 50.854223 & 7.076385 & 7.162433 & 5.680042 & 7.597738 & 16.799287 & 8.168466 & 23.049238 & 6.217165 & 8.85919 & 5.98005 & 6.220977 & 5.507643 & 8.019513 & 11.811277 & 7.590282 \\
    & $E_\mathrm{O}^\mathrm{med}$ & 4.891983 & 5.813537 & 5.954739 & 4.655276 & 5.488949 & 6.263676 & 33.380847 & 7.772206 & 5.685236 & 49.412944 & 6.950258 & 6.479874 & 5.323128 & 7.209721 & 15.690739 & 7.836117 & 22.378667 & 6.119478 & 8.385309 & 5.547806 & 5.850552 & 5.299148 & 7.574797 & 11.996856 & 7.167634 \\
    & $E_\mathrm{O}^\mathrm{std}$ & 0.279153 & 0.384793 & 0.337127 & 0.280523 & 0.296554 & 0.364466 & 2.88465 & 0.409637 & 0.347122 & 1.292708 & 0.396281 & 0.403734 & 0.325988 & 0.389981 & 0.789146 & 0.406186 & 0.725185 & 0.35717 & 0.398588 & 0.310507 & 0.332691 & 0.321249 & 0.385675 & 0.513163 & 0.348818 \\
    & $E_\mathrm{BBC}^\mathrm{avg}$ & 3.83998 & 3.758785 & 4.72131 & 2.662871 & 3.640138 & 5.376508 & 35.463433 & 5.575918 & 3.676049 & 50.645706 & 5.137198 & 6.17226 & 4.908205 & 5.258787 & 14.416551 & 6.684433 & 22.247523 & 4.522085 & 7.239723 & 3.95819 & 4.97029 & 3.491991 & 6.605975 & 9.189634 & 5.790274 \\
    & $E_\mathrm{BBC}^\mathrm{med}$ & 3.687449 & 3.704441 & 4.619835 & 2.592074 & 3.282285 & 5.110269 & 30.25958 & 5.157557 & 3.333755 & 49.201898 & 4.893599 & 5.570751 & 4.520831 & 5.099914 & 13.603499 & 6.347707 & 21.3493 & 4.429801 & 7.143247 & 3.79934 & 4.592891 & 3.31541 & 5.997683 & 9.581055 & 5.436204 \\
    & $E_\mathrm{BBC}^\mathrm{std}$ & 0.200162 & 0.275784 & 0.287592 & 0.18128 & 0.214463 & 0.317461 & 2.971732 & 0.322089 & 0.255328 & 1.297412 & 0.327187 & 0.37077 & 0.308292 & 0.292224 & 0.773887 & 0.363048 & 0.739171 & 0.278083 & 0.360257 & 0.237076 & 0.284057 & 0.230992 & 0.346641 & 0.459693 & 0.316927 \\
      \hline
    \multirow{6}[0]{*}{25} & $E_\mathrm{O}^\mathrm{avg}$ & 4.078403 & 5.774284 & 5.182098 & 5.046066 & 4.619734 & 10.760324 & 40.436873 & 5.990182 & 5.910632 & 46.637053 & 10.36862 & 5.799561 & 3.890338 & 6.531268 & 19.170333 & 7.889456 & 26.146364 & 6.015266 & 8.103984 & 4.843508 & 4.824584 & 4.63869 & 7.711545 & 12.559944 & 5.890261 \\
    & $E_\mathrm{O}^\mathrm{med}$ & 3.947814 & 4.682853 & 4.793935 & 4.171959 & 4.433424 & 9.958962 & 42.45874 & 5.662901 & 5.723091 & 46.451811 & 9.304376 & 5.51753 & 3.691323 & 6.032344 & 18.600762 & 7.825783 & 26.648521 & 5.933501 & 7.672882 & 4.623775 & 4.674247 & 4.520197 & 7.507207 & 12.559721 & 5.507783 \\
    & $E_\mathrm{O}^\mathrm{std}$ & 0.238925 & 0.587756 & 0.301748 & 0.382501 & 0.271254 & 0.800366 & 2.26323 & 0.285034 & 0.303029 & 0.891279 & 0.634511 & 0.281921 & 0.209864 & 0.317004 & 0.855751 & 0.338923 & 0.766223 & 0.302311 & 0.317131 & 0.216325 & 0.24811 & 0.269617 & 0.341541 & 0.428722 & 0.232588 \\
    & $E_\mathrm{BBC}^\mathrm{avg}$ & 3.135131 & 4.245119 & 4.10966 & 3.421183 & 3.414164 & 9.663474 & 38.0104 & 4.103635 & 4.002726 & 46.345424 & 8.632429 & 4.728421 & 3.002642 & 4.6182 & 17.250969 & 6.052937 & 25.265887 & 4.224709 & 6.450708 & 3.727292 & 3.691834 & 3.464557 & 5.862637 & 9.976237 & 4.215947 \\
    & $E_\mathrm{BBC}^\mathrm{med}$ & 2.800137 & 3.604413 & 3.853352 & 2.847745 & 3.178394 & 8.645056 & 40.304905 & 3.791404 & 3.802165 & 46.302824 & 7.797221 & 4.461402 & 2.71819 & 4.171489 & 16.508949 & 5.782914 & 25.755481 & 4.246699 & 6.26323 & 3.508624 & 3.514498 & 3.325804 & 5.896379 & 9.559468 & 4.068204 \\
    & $E_\mathrm{BBC}^\mathrm{std}$ & 0.197214 & 0.454714 & 0.249527 & 0.296946 & 0.198047 & 0.798439 & 2.359074 & 0.194435 & 0.224669 & 0.890702 & 0.58854 & 0.237777 & 0.168603 & 0.236547 & 0.802964 & 0.272567 & 0.76607 & 0.233089 & 0.258505 & 0.185345 & 0.197161 & 0.208185 & 0.273243 & 0.375031 & 0.188071 \\
      \hline
    \multirow{6}[0]{*}{50} & $E_\mathrm{O}^\mathrm{avg}$ & 4.199464 & 4.723996 & 4.646755 & 5.328964 & 4.343066 & 11.049719 & 35.337385 & 4.96099 & 5.983577 & 46.609801 & 10.848115 & 6.320416 & 3.326026 & 5.826521 & 19.702562 & 7.403735 & 23.400784 & 5.65382 & 7.346276 & 4.70901 & 4.414316 & 4.447518 & 7.358356 & 12.525741 & 5.513642 \\
    & $E_\mathrm{O}^\mathrm{med}$ & 4.453838 & 4.763492 & 4.529226 & 4.929194 & 4.027858 & 10.324134 & 36.410904 & 4.853719 & 6.200398 & 45.756363 & 10.479451 & 5.90261 & 3.165678 & 6.054243 & 20.053 & 7.421184 & 23.968394 & 5.923401 & 7.351914 & 4.726675 & 4.306853 & 4.458327 & 7.455512 & 12.468697 & 5.450241 \\
    & $E_\mathrm{O}^\mathrm{std}$ & 0.290061 & 0.326044 & 0.221913 & 0.41455 & 0.256495 & 0.668899 & 2.160774 & 0.260893 & 0.379051 & 1.121469 & 0.6534 & 0.349498 & 0.188924 & 0.236631 & 0.8221 & 0.326453 & 0.72157 & 0.289044 & 0.24706 & 0.187357 & 0.284126 & 0.2736 & 0.329389 & 0.416199 & 0.173817 \\
    & $E_\mathrm{BBC}^\mathrm{avg}$ & 3.255193 & 3.615984 & 3.717092 & 4.017348 & 3.180156 & 10.008797 & 32.557785 & 3.460917 & 4.421475 & 46.313135 & 9.263762 & 5.246289 & 2.396605 & 4.123805 & 17.923053 & 5.608418 & 22.494355 & 3.90353 & 5.51403 & 3.572909 & 3.237428 & 3.288414 & 5.595332 & 9.737858 & 3.828889 \\
    & $E_\mathrm{BBC}^\mathrm{med}$ & 3.310295 & 3.755224 & 3.624177 & 3.717186 & 2.983567 & 9.293215 & 32.156423 & 3.345729 & 4.49338 & 45.454091 & 8.734466 & 4.866918 & 2.405959 & 4.235595 & 17.948825 & 5.332289 & 23.337576 & 3.896853 & 5.375247 & 3.413408 & 3.161034 & 3.200045 & 5.626498 & 9.717429 & 3.741931 \\
    & $E_\mathrm{BBC}^\mathrm{std}$ & 0.240433 & 0.272959 & 0.173091 & 0.336766 & 0.1685 & 0.637148 & 2.180623 & 0.152701 & 0.302724 & 1.125571 & 0.600682 & 0.298767 & 0.134754 & 0.156153 & 0.773006 & 0.246891 & 0.707657 & 0.196496 & 0.187161 & 0.151315 & 0.200962 & 0.194178 & 0.23935 & 0.322404 & 0.12099 \\
      \hline
    \multirow{6}[0]{*}{100} & $E_\mathrm{O}^\mathrm{avg}$ & 4.366919 & 6.129979 & 5.109057 & 6.086914 & 4.620188 & 13.069763 & 37.660771 & 4.517186 & 6.346963 & 47.25588 & 12.16768 & 6.93957 & 2.983978 & 5.795045 & 20.393792 & 7.198874 & 24.245319 & 5.832677 & 6.63639 & 5.09088 & 4.658611 & 4.900801 & 7.240282 & 13.152027 & 5.17522 \\
    & $E_\mathrm{O}^\mathrm{med}$ & 4.331512 & 5.232082 & 4.903406 & 5.822958 & 4.684227 & 12.902987 & 36.824781 & 4.577678 & 6.110871 & 47.224328 & 11.739947 & 6.112198 & 3.028512 & 5.907096 & 19.914226 & 7.057504 & 24.494086 & 6.047963 & 6.643434 & 4.938985 & 4.542633 & 4.744797 & 7.24306 & 13.471561 & 5.272757 \\
    & $E_\mathrm{O}^\mathrm{std}$ & 0.200894 & 0.464284 & 0.176961 & 0.315784 & 0.181333 & 0.592083 & 2.497756 & 0.148361 & 0.327622 & 1.15586 & 0.482045 & 0.351426 & 0.106135 & 0.14217 & 0.547887 & 0.19934 & 0.559278 & 0.203331 & 0.133254 & 0.143672 & 0.222658 & 0.225832 & 0.198905 & 0.269683 & 0.114586 \\
    & $E_\mathrm{BBC}^\mathrm{avg}$ & 3.466779 & 4.820443 & 4.016903 & 4.860445 & 3.464279 & 12.021397 & 35.113133 & 3.176135 & 4.955944 & 46.942448 & 10.583807 & 5.810155 & 2.051397 & 4.063478 & 18.667963 & 5.41935 & 23.295769 & 4.136291 & 4.60667 & 3.954857 & 3.395401 & 3.68464 & 5.475998 & 9.625924 & 3.516935 \\
    & $E_\mathrm{BBC}^\mathrm{med}$ & 3.580303 & 4.008706 & 3.828292 & 4.601501 & 3.511971 & 11.505672 & 32.811852 & 3.175859 & 4.648323 & 46.94156 & 10.292957 & 4.873515 & 2.054739 & 4.097966 & 18.306492 & 5.543078 & 23.484869 & 4.074637 & 4.689083 & 3.774418 & 3.29339 & 3.56119 & 5.424779 & 9.872627 & 3.482706 \\
    & $E_\mathrm{BBC}^\mathrm{std}$ & 0.162148 & 0.380047 & 0.14915 & 0.276155 & 0.12015 & 0.588404 & 2.605332 & 0.092934 & 0.262857 & 1.162753 & 0.453231 & 0.311748 & 0.077071 & 0.118922 & 0.520758 & 0.145331 & 0.553551 & 0.129779 & 0.099786 & 0.136054 & 0.151361 & 0.149016 & 0.152101 & 0.232881 & 0.082889 \\
      \hline
      \hline
    \end{tabular}}
\end{table}

\begin{table}[!t]
\color{black}
\centering
  \caption{\color{black}
Performance statistics of algorithms on the GMPB benchmark, evaluated across different instances with 5, 10, 25, 50, and 100 promising regions. The table shows the average, median, and standard deviation of the offline error and the average best error before environmental changes, based on 31 runs for each scenario. }
  \label{tab:GMPBresults}
\Rotatebox{90}{
\tiny
    \begin{tabular}{c|c|ccccccccccccccccccccccccc}
      \hline
      \hline
    P     & I & ACFPSO & AMPDE & AMPPSO & AMSO & AmQSO & CDE & CESO & CPSO & CPSOR & DSPSO & DynDE & DynPopDE & FTMPSO & HmSO & IDSPSO & ImQSO & RPSO & SPSO$_\mathrm{AD+AP}$ & TMIPSO & mCMAES & mDE & mPSO & mQSO & mjDE & psfNBC \\
      \hline
      \multirow{6}[0]{*}{5} & $E_\mathrm{O}^\mathrm{avg}$ & 2.097182 & 2.517297 & 1.727482 & 2.570194 & 2.458395 & 27.153524 & 14.071055 & 4.628525 & 4.866031 & 90.581681 & 31.501787 & 23.495933 & 3.284523 & 3.976039 & 22.519032 & 2.687927 & 14.292336 & 1.884522 & 5.106298 & 1.739823 & 2.450614 & 2.364962 & 2.688841 & 2.975802 & 2.145919 \\
      & $E_\mathrm{O}^\mathrm{med}$ & 1.959947 & 2.497789 & 1.406037 & 1.996835 & 2.44445 & 24.618098 & 13.071048 & 4.684182 & 3.933908 & 89.879997 & 30.034982 & 21.006773 & 3.078246 & 3.454261 & 22.961301 & 2.571852 & 15.111865 & 1.771624 & 4.881553 & 1.46649 & 2.36987 & 2.051899 & 2.614677 & 2.232701 & 1.895157 \\
      & $E_\mathrm{O}^\mathrm{std}$ & 0.138679 & 0.052276 & 0.156932 & 0.264646 & 0.102394 & 2.211577 & 0.96102 & 0.201599 & 0.56297 & 1.887641 & 2.489402 & 1.73956 & 0.140938 & 0.341038 & 1.238738 & 0.117072 & 0.578488 & 0.111075 & 0.168521 & 0.192852 & 0.104532 & 0.117521 & 0.12286 & 0.413427 & 0.127417 \\
      & $E_\mathrm{BBC}^\mathrm{avg}$ & 1.363676 & 1.868222 & 1.042016 & 1.663592 & 1.423356 & 24.832629 & 13.639149 & 2.969647 & 3.813071 & 90.067524 & 28.806258 & 20.022856 & 1.939775 & 2.984042 & 21.022454 & 1.828218 & 13.918387 & 1.421132 & 3.906597 & 1.34259 & 1.696811 & 1.424301 & 1.775053 & 2.207571 & 1.156617 \\
      & $E_\mathrm{BBC}^\mathrm{med}$ & 1.22808 & 1.923847 & 0.698694 & 1.17077 & 1.337354 & 22.609582 & 12.720784 & 2.873251 & 3.018368 & 89.333331 & 27.48213 & 17.129851 & 1.83038 & 2.578401 & 21.443862 & 1.689905 & 14.745101 & 1.315349 & 3.626746 & 1.076459 & 1.67201 & 1.203642 & 1.588237 & 1.506716 & 0.837696 \\
      & $E_\mathrm{BBC}^\mathrm{std}$ & 0.121085 & 0.05383 & 0.144944 & 0.232903 & 0.077195 & 2.069581 & 0.966877 & 0.166599 & 0.521997 & 1.893007 & 2.295129 & 1.529021 & 0.112093 & 0.318865 & 1.16188 & 0.099832 & 0.575537 & 0.102783 & 0.152701 & 0.185351 & 0.08947 & 0.103012 & 0.098211 & 0.370247 & 0.12843 \\
        \hline
      \multirow{6}[0]{*}{10} & $E_\mathrm{O}^\mathrm{avg}$ & 2.33858 & 2.347558 & 1.531063 & 3.038471 & 2.301531 & 25.365231 & 16.716149 & 4.408601 & 3.737657 & 81.599094 & 28.038519 & 20.23552 & 2.92807 & 4.453806 & 19.394413 & 2.346602 & 17.644526 & 1.677764 & 5.453112 & 1.838854 & 2.331231 & 2.187059 & 2.21308 & 4.045937 & 2.700425 \\
      & $E_\mathrm{O}^\mathrm{med}$ & 2.228527 & 2.200889 & 1.497271 & 2.351599 & 2.230225 & 21.583371 & 16.511306 & 4.373452 & 3.064046 & 82.782563 & 28.407116 & 20.975067 & 2.810174 & 4.037342 & 18.786857 & 2.368827 & 17.524665 & 1.557093 & 5.346371 & 1.633305 & 2.355167 & 2.168768 & 2.197943 & 3.819046 & 2.546861 \\
      & $E_\mathrm{O}^\mathrm{std}$ & 0.139869 & 0.125011 & 0.07214 & 0.569458 & 0.063123 & 1.792615 & 0.690314 & 0.158238 & 0.32635 & 1.912445 & 1.790819 & 1.181972 & 0.09613 & 0.323913 & 0.832654 & 0.092477 & 0.726748 & 0.102309 & 0.17004 & 0.118209 & 0.071842 & 0.066667 & 0.083572 & 0.293745 & 0.127047 \\
      & $E_\mathrm{BBC}^\mathrm{avg}$ & 1.621291 & 1.673271 & 0.870673 & 2.268498 & 1.341281 & 23.479959 & 16.285359 & 2.922825 & 2.903695 & 81.168626 & 25.836471 & 17.622997 & 1.74059 & 3.575586 & 18.017963 & 1.666513 & 17.271088 & 1.205739 & 4.201108 & 1.331285 & 1.558514 & 1.261382 & 1.517623 & 3.226812 & 1.647089 \\
      & $E_\mathrm{BBC}^\mathrm{med}$ & 1.47038 & 1.534786 & 0.835173 & 1.591337 & 1.323839 & 19.851216 & 16.185907 & 2.9002 & 2.363521 & 82.311659 & 26.306706 & 18.268127 & 1.641787 & 3.302556 & 17.532518 & 1.668061 & 17.211805 & 1.082454 & 4.071264 & 1.200528 & 1.460936 & 1.236296 & 1.508126 & 3.190941 & 1.430334 \\
      & $E_\mathrm{BBC}^\mathrm{std}$ & 0.133109 & 0.123819 & 0.063771 & 0.535013 & 0.054917 & 1.679059 & 0.696375 & 0.148911 & 0.302506 & 1.922756 & 1.677993 & 1.050165 & 0.090606 & 0.302725 & 0.797919 & 0.088332 & 0.728543 & 0.098905 & 0.162266 & 0.115984 & 0.06279 & 0.058078 & 0.077821 & 0.283425 & 0.116424 \\
        \hline
      \multirow{6}[0]{*}{25} & $E_\mathrm{O}^\mathrm{avg}$ & 2.677841 & 2.812229 & 2.218013 & 3.163067 & 2.558616 & 24.915867 & 18.493456 & 4.381953 & 4.382943 & 69.333233 & 26.989616 & 21.84394 & 3.058818 & 4.493123 & 17.540192 & 3.586907 & 18.288899 & 1.944114 & 6.244285 & 2.579684 & 2.823332 & 2.552841 & 3.495267 & 8.568641 & 4.192272 \\
      & $E_\mathrm{O}^\mathrm{med}$ & 2.691559 & 2.479583 & 2.094673 & 3.048652 & 2.559482 & 23.422373 & 18.889597 & 4.412245 & 4.168725 & 69.575191 & 28.080262 & 21.139105 & 3.092114 & 4.390062 & 17.559308 & 3.470012 & 19.372805 & 1.948865 & 6.080617 & 2.597074 & 2.773471 & 2.548504 & 3.400905 & 6.627917 & 4.211646 \\
      & $E_\mathrm{O}^\mathrm{std}$ & 0.080075 & 0.17342 & 0.085853 & 0.172444 & 0.035914 & 1.375731 & 0.739092 & 0.09215 & 0.234071 & 1.166746 & 1.4615 & 1.168805 & 0.078928 & 0.17728 & 0.702832 & 0.114814 & 0.653432 & 0.044763 & 0.139295 & 0.078302 & 0.069735 & 0.053882 & 0.122924 & 0.889956 & 0.11858 \\
      & $E_\mathrm{BBC}^\mathrm{avg}$ & 1.975908 & 2.119472 & 1.533106 & 2.455738 & 1.615595 & 23.274689 & 18.060105 & 3.027247 & 3.662339 & 68.972415 & 25.119154 & 19.234897 & 1.921097 & 3.640132 & 16.346258 & 2.877646 & 17.890118 & 1.322355 & 4.908325 & 1.968192 & 1.949477 & 1.610712 & 2.809117 & 7.576321 & 3.166439 \\
      & $E_\mathrm{BBC}^\mathrm{med}$ & 2.006007 & 1.785288 & 1.462529 & 2.366514 & 1.602482 & 21.955746 & 18.35113 & 2.945645 & 3.526112 & 69.280144 & 26.387903 & 18.306467 & 1.978352 & 3.663758 & 16.26766 & 2.864276 & 18.902691 & 1.342299 & 4.752063 & 1.988532 & 1.917553 & 1.605203 & 2.714866 & 5.805043 & 3.192767 \\
      & $E_\mathrm{BBC}^\mathrm{std}$ & 0.072884 & 0.174828 & 0.082383 & 0.163257 & 0.032081 & 1.314026 & 0.73526 & 0.083891 & 0.226948 & 1.175135 & 1.34537 & 1.05866 & 0.070949 & 0.16024 & 0.663353 & 0.104725 & 0.650255 & 0.040605 & 0.122068 & 0.07803 & 0.063061 & 0.048417 & 0.117629 & 0.829726 & 0.11716 \\
        \hline
      \multirow{6}[0]{*}{50} & $E_\mathrm{O}^\mathrm{avg}$ & 2.770733 & 2.923618 & 2.556958 & 3.555832 & 2.867675 & 22.362065 & 19.245183 & 4.200237 & 4.210187 & 62.612286 & 24.14046 & 17.239833 & 3.03186 & 4.624178 & 17.753878 & 3.694566 & 16.72091 & 2.180439 & 6.335511 & 2.905464 & 2.949292 & 2.754594 & 3.729379 & 10.550755 & 4.398332 \\
      & $E_\mathrm{O}^\mathrm{med}$ & 2.770779 & 2.681298 & 2.527926 & 3.401749 & 2.791135 & 21.429124 & 19.376928 & 4.239029 & 4.207926 & 63.583568 & 24.949264 & 17.91999 & 3.040055 & 4.300024 & 17.601448 & 3.730245 & 16.33129 & 2.214228 & 6.165418 & 2.82966 & 2.995211 & 2.700796 & 3.705549 & 9.351387 & 4.468022 \\
      & $E_\mathrm{O}^\mathrm{std}$ & 0.086301 & 0.147123 & 0.068063 & 0.132149 & 0.057773 & 1.088426 & 0.791078 & 0.072027 & 0.155546 & 0.872641 & 1.04568 & 0.731263 & 0.083587 & 0.174681 & 0.560783 & 0.086852 & 0.46521 & 0.051281 & 0.101752 & 0.086701 & 0.052618 & 0.057073 & 0.085089 & 0.956011 & 0.095157 \\
      & $E_\mathrm{BBC}^\mathrm{avg}$ & 2.100651 & 2.243136 & 1.865823 & 2.860948 & 1.92441 & 20.949668 & 18.802765 & 2.917658 & 3.515755 & 62.345819 & 22.526022 & 14.978382 & 1.954034 & 3.77283 & 16.513789 & 3.022346 & 16.302981 & 1.525198 & 4.982762 & 2.267504 & 2.022651 & 1.844062 & 3.057399 & 9.453898 & 3.361239 \\
      & $E_\mathrm{BBC}^\mathrm{med}$ & 2.027316 & 1.937253 & 1.814775 & 2.72966 & 1.935982 & 20.00242 & 19.033791 & 2.95065 & 3.483645 & 63.286452 & 23.016041 & 15.021247 & 1.979431 & 3.515533 & 16.363897 & 3.058267 & 15.959644 & 1.550414 & 4.798355 & 2.182481 & 2.018071 & 1.788966 & 3.05421 & 8.526454 & 3.420703 \\
      & $E_\mathrm{BBC}^\mathrm{std}$ & 0.078787 & 0.148979 & 0.06682 & 0.126769 & 0.048736 & 1.042661 & 0.797226 & 0.064803 & 0.148034 & 0.871659 & 0.982771 & 0.660026 & 0.074165 & 0.155725 & 0.54026 & 0.08123 & 0.461915 & 0.048014 & 0.091588 & 0.081792 & 0.047642 & 0.050897 & 0.079543 & 0.890388 & 0.093108 \\
        \hline
      \multirow{6}[0]{*}{100} & $E_\mathrm{O}^\mathrm{avg}$ & 2.763346 & 3.336214 & 2.698681 & 3.670947 & 3.020507 & 21.645177 & 18.896305 & 4.142338 & 4.097674 & 53.917177 & 22.168973 & 17.384061 & 3.016697 & 4.109834 & 16.787474 & 3.733904 & 15.218057 & 2.404156 & 6.633355 & 3.112902 & 3.11709 & 3.005878 & 3.786216 & 12.120551 & 4.272754 \\
      & $E_\mathrm{O}^\mathrm{med}$ & 2.774374 & 2.910335 & 2.672255 & 3.822699 & 2.949968 & 22.33925 & 18.938253 & 4.124489 & 3.963906 & 53.599898 & 21.290186 & 16.885388 & 2.905584 & 4.110188 & 16.125259 & 3.727187 & 15.716285 & 2.355466 & 6.747017 & 2.989435 & 3.123627 & 3.018566 & 3.777924 & 10.210886 & 4.203261 \\
      & $E_\mathrm{O}^\mathrm{std}$ & 0.068417 & 0.160401 & 0.051622 & 0.114716 & 0.057205 & 1.100604 & 0.701056 & 0.059149 & 0.097326 & 0.715884 & 1.17868 & 0.742408 & 0.065851 & 0.089568 & 0.511772 & 0.067117 & 0.514677 & 0.041209 & 0.121471 & 0.100732 & 0.048018 & 0.048654 & 0.064825 & 1.011368 & 0.090305 \\
      & $E_\mathrm{BBC}^\mathrm{avg}$ & 2.154408 & 2.66916 & 2.019371 & 3.020897 & 2.094268 & 20.390626 & 18.450945 & 2.875428 & 3.443643 & 53.651272 & 20.662576 & 15.249135 & 1.954433 & 3.342772 & 15.64327 & 3.055749 & 14.736768 & 1.766995 & 5.291087 & 2.482326 & 2.196776 & 2.102483 & 3.134485 & 10.861061 & 3.245077 \\
      & $E_\mathrm{BBC}^\mathrm{med}$ & 2.148496 & 2.250015 & 2.022896 & 3.108537 & 2.053599 & 21.197014 & 18.082062 & 2.854534 & 3.33284 & 53.331821 & 19.555749 & 15.118557 & 1.892324 & 3.421063 & 15.069094 & 3.106871 & 15.203491 & 1.767914 & 5.367111 & 2.344711 & 2.237605 & 2.054606 & 3.116602 & 9.15107 & 3.269911 \\
      & $E_\mathrm{BBC}^\mathrm{std}$ & 0.062857 & 0.161015 & 0.047511 & 0.108858 & 0.047363 & 1.049406 & 0.698435 & 0.051173 & 0.09555 & 0.717325 & 1.118523 & 0.647179 & 0.060633 & 0.083826 & 0.493335 & 0.058463 & 0.509057 & 0.037399 & 0.100404 & 0.096673 & 0.039456 & 0.043971 & 0.056869 & 0.941948 & 0.088681 \\
      \hline
      \hline
    \end{tabular}%
  }
\end{table}

\clearpage

\newpage

\color{black}

\renewcommand{\thetable}{A2-\arabic{table}} 
\setcounter{table}{0}

\renewcommand{\thefigure}{A2-\arabic{figure}} 
\setcounter{figure}{0}

\renewcommand{\thesection}{A2-\arabic{section}} 
\setcounter{section}{0}

\part*{\LARGE Appendix 2: User Manual}

{This user manual provides a comprehensive guide for running, configuring, and extending EDOLAB, an open-source MATLAB platform for evolutionary dynamic optimization algorithms (EDOAs). EDOLAB includes two key modules: the \textit{Education} module, which visualizes algorithm behavior over time, and the \textit{Experimentation} module, designed for conducting experiments and comparing algorithms. The platform supports both GUI and non-GUI modes, offering flexibility for users. Additionally, instructions are provided for running EDOLAB in Octave, an open-source alternative to MATLAB. To begin, clone the EDOLAB project from the GitHub repository: [\url{https://github.com/Danial-Yazdani/EDOLAB-MATLAB}].}

\section{Architecture}

EDOLAB is a function-based software implemented in MATLAB. 
The \emph{MATLAB App Designer} was used to develop the GUI for EDOLAB. 
The software can be operated either with or without the GUI.

The root directory of EDOLAB includes the following:

\begin{itemize} 
\item A $\mathtt{.MLAPP}$ file, which is the GUI developed using MATLAB App Designer. 
\item Two $\mathtt{.m}$ files: \begin{inparaenum} \item $\mathtt{RunWithGUI.m}$--the exported GUI $\mathtt{.m}$ file, and \item $\mathtt{RunWithoutGUI.m}$--a function for using EDOLAB without the GUI. \end{inparaenum} 
\item Five folders: \begin{itemize} 
\item \textbf{Algorithm}: This folder contains several sub-folders, each corresponding to an EDOA listed in Table~\ref{tab:algorithms}.
Each EDOA sub-folder generally includes several $\mathtt{.m}$ files: 
\begin{inparaenum} 
\item $\mathtt{main\_EDOA.m}$--the main file that invokes and controls other EDOA functions, 
\item $\mathtt{SubPopulationGenerator\_EDOA.m}$--a sub-population generator function that generates the sub-populations for the optimization component,
\item $\mathtt{IterativeComponents\_EDOA.m}$--a function containing the EDOA components that are executed every iteration, or when certain conditions are met, and 
\item $\mathtt{ChangeReaction\_EDOA.m}$--a function that includes the change reaction components of the EDOA.
\end{inparaenum} 
\item \textbf{Benchmark}: This folder contains a sub-folder for each benchmark generator included in EDOLAB. Each benchmark sub-folder includes two $\mathtt{.m}$ files: 
\begin{inparaenum} 
\item $\mathtt{BenchmarkGenerator\_Benchmark.m}$--responsible for setting up the benchmark and generating environments, and
\item $\mathtt{fitness\_Benchmark.m}$--includes the baseline function of the benchmark for calculating function values. 
\end{inparaenum} 
These functions are called by three $\mathtt{.m}$ files located in the $\mathtt{Benchmark}$ folder: 
\begin{inparaenum} 
\item $\mathtt{BenchmarkGenerator.m}$--invokes the initializer and generator of the benchmark problem selected by the experimenter, 
\item $\mathtt{fitness.m}$--calls the related benchmark's baseline function for calculating fitness values, manages benchmark parameters, counters, and flags, and gathers the information needed to calculate performance indicators, and \item $\mathtt{EnvironmentVisualization.m}$--an environment visualization function responsible for depicting the problem landscape in the educational module. 
\end{inparaenum} 
\item \textbf{Results}: For each experiment, EDOLAB generates an Excel file (if selected by the user) that contains the results, statistics, and experiment settings. 
These output Excel files are stored in this folder. 
\item \textbf{Utility}: This folder includes various utility functions, such as those for generating output files and figures, which are located in the $\mathtt{Output}$ sub-folder. 
\item {\textbf{Octave\_compatibility}: This folder contains updated versions of key files modified for compatibility with Octave, including $\mathtt{RunWithoutGUI.m}$ and several others. Users wishing to run EDOLAB in Octave should replace the corresponding files in the main EDOLAB directory with those provided in this folder.}
\end{itemize} 
\end{itemize}

\begin{figure}[!t]
\centering
\includegraphics[width=0.999\linewidth]{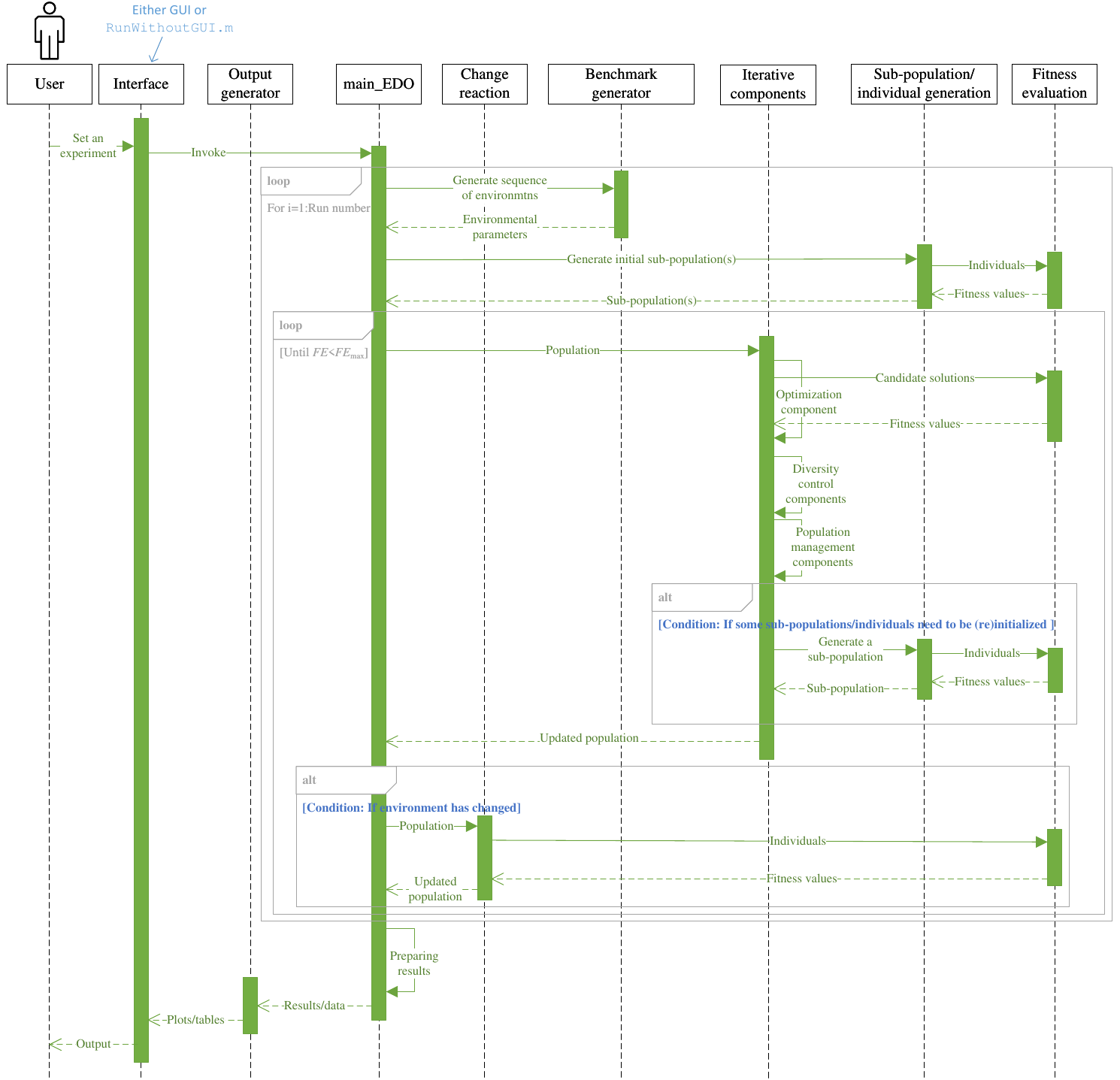}
\caption{A general sequence diagram of running an EDOA in EDOLAB.}
\label{fig:SequenceDiagram}
\end{figure}

Figure~\ref{fig:SequenceDiagram} illustrates a general sequence diagram for running an EDOA in EDOLAB, which demonstrates how the platform operates. 
First, the user sets up an experiment using either the GUI or the $\mathtt{RunWithoutGUI.m}$ and initiates the run.
The interface then invokes the main function of the selected algorithm (for example, $\mathtt{main\_AmQSO.m}$). 
At the start of the main function, the benchmark generator function ($\mathtt{BenchmarkGenerator.m}$) is called. 
This function is responsible for initializing the benchmark and generating a sequence of environments based on the parameters defined by the user.

In EDOLAB's experimentation module, identical random streams are used when initializing problem instances and generating environmental changes in $\mathtt{BenchmarkGenerator.m}$. 
As a result, with the same parameter settings, the same problem instance sequence (from the first environment to the last) is generated for all comparison algorithms. 
Using different random seeds in experiments can produce problem instances with varying characteristics and difficulty levels~\cite{yazdani2021DOPsurveyPartB}, potentially leading to biased comparisons. 
In EDOLAB, we have addressed this issue by controlling the random streams.
After generating the sequence of environments, the initial sub-population(s) or individuals are generated by the sub-population/individual generation function (for example, $\mathtt{SubPopulationGenerator\_AmQSO.m}$).

Afterward, the main loop of the EDOA is executed. 
In each iteration, the iterative components of the EDOA, such as the optimizer (e.g., PSO or DE), diversity control, and population management~\cite{yazdani2020adaptive}, are executed by calling the iterative components function (for example, $\mathtt{IterativeComponents\_AmQSO.m}$). 
In many EDOAs with adaptive sub-population numbers and/or population sizes, new individuals or sub-populations are generated when certain conditions are met~\cite{yazdani2021DOPsurveyPartA}. 
Additionally, some diversity and population management components may require the reinitialization of certain sub-populations or individuals. 
Therefore, if any sub-populations or individuals need to be (re)initialized during an iteration, the sub-population/individual generation function is called. 
The updated population is then returned to the main EDOA function.

At the end of each iteration, if the environment has changed, the change reaction components are called (for example, $\mathtt{ChangeReaction\_AmQSO.m}$). 
The main loop of the EDOA continues until the number of function evaluations ($FE$) reaches its predefined maximum value ($FE_\mathrm{max}$). 
This procedure is repeated for the specified number of runs ($\mathtt{RunNumber}$). 
Afterward, the results are processed, including the calculation of performance indicators. 
The results, along with any collected data, are then sent to output generator functions responsible for creating output plots, tables, and files. 
Finally, the output tables and figures are returned to the interface.

\section{Running}
\label{sec:running}

As previously mentioned, EDOLAB can be operated either with or without a GUI. 
In the following, we describe both methods of use.

\subsection{Using EDOLAB via GUI}
\label{sec:sec:sec:GUI}

The GUI for EDOLAB is developed using MATLAB App Designer and can be accessed by executing either $\mathtt{GUI.MLAPP}$ or $\mathtt{RunWithGUI.m}$ from the root directory of EDOLAB.
Note that the GUI is designed for MATLAB R2020b and is not backward compatible. 
Therefore, to use EDOLAB with the GUI, the user must have MATLAB R2020b or a newer version. 
Users with older MATLAB versions can still use EDOLAB by running $\mathtt{RunWithoutGUI.m}$ (see Section~\ref{sec:sec:RunWithoutGUI}).
EDOLAB's GUI contains two modules---\emph{Experimentation} and \emph{Education}---which are explained below.

\subsubsection{Experimentation module}
\label{sec:sec:exprementationModule}
\begin{figure}[!t]
\centering
\includegraphics[width=0.9\linewidth]{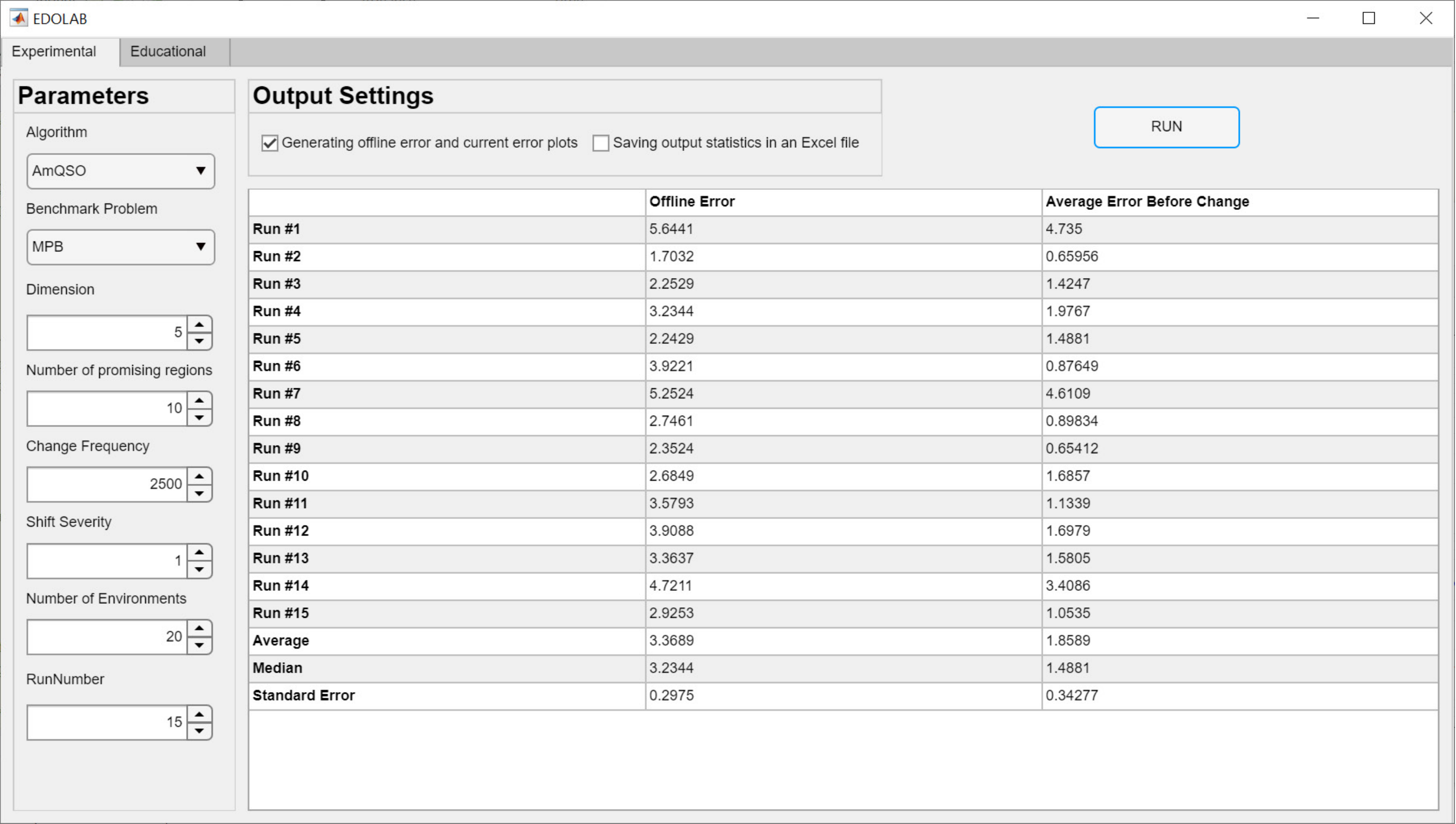}
\caption{The experimentation module of EDOLAB.}
\label{fig:ExperimentationModul}
\end{figure}

The experimentation module is designed for conducting experiments. 
Figure~\ref{fig:ExperimentationModul} shows the interface of this module, where users can select the algorithm (EDOA) and the benchmark generator. 
Additionally, users can configure the parameters of the benchmark generator to generate the desired problem instance.
Note that EDOLAB's GUI does not provide an option to adjust the parameter settings of the EDOAs. 
This is because EDOAs typically have numerous parameters, which vary across algorithms depending on their structural components. 
Adding a feature to modify these parameters in the GUI would significantly increase complexity and make the interface harder to use and more confusing. 
Therefore, in EDOLAB, the parameters for each EDOA are preset based on the recommended values from their original references. 
Our evaluations show that these settings yield the best performance for the EDOAs. 
For users interested in performing sensitivity analysis on EDOA parameters, adjustments can be made directly in the source code.

As illustrated in Figure~\ref{fig:ExperimentationModul}, users can configure the number of runs and several key benchmark parameters, such as dimension, number of promising regions, change frequency, shift severity, and the number of environments---these parameters are common between MPB, GDBG, GMPB, and FPs. 
The type and recommended values for these parameters are provided in Table~\ref{tab:BenchmarkParameters}. 
In most studies, only the dimension, number of promising regions, change frequency, and shift severity are modified to generate different problem instances.
Finally, users need to configure the ``output settings.'' 
Using two checkboxes, they can choose whether to generate a figure with offline error and current error plots, and/or an Excel file containing the experiment results and statistics.

Once the experiment configuration is complete, the experiment can be started by pressing the $\mathtt{RUN}$ button in the top-right corner of the interface. 
The duration of the experiment depends on the complexity of the chosen EDOA and the configured problem instance, and it may take a significant amount of time to finish. 
It is worth noting that due to the complexity of EDOAs and dynamic benchmark generators, runs in this field generally take longer than those in other sub-fields of evolutionary computation, such as evolutionary static optimization or evolutionary multi-objective optimization with similar problem dimensionalities.

\begin{figure}[!t]
\centering
\fbox{\includegraphics[width=0.6\linewidth]{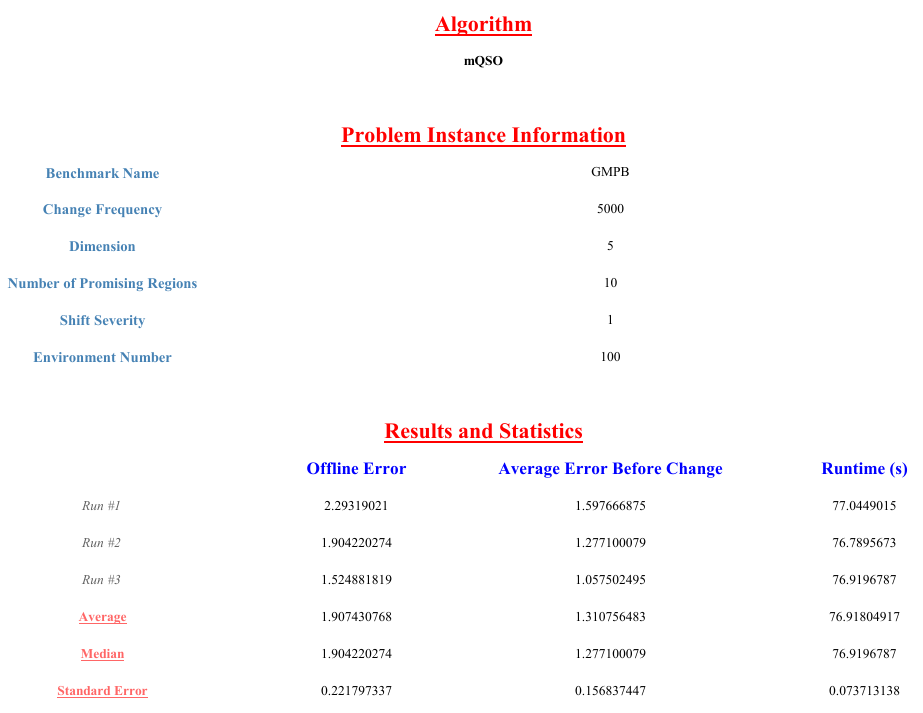}}
\caption{An Excel output table generated by EDOLAB's $\mathtt{OutputExcel.m}$ function.}
\label{fig:Excel}
\end{figure}

\begin{figure}[!t]
\centering
\includegraphics[width=0.8\linewidth]{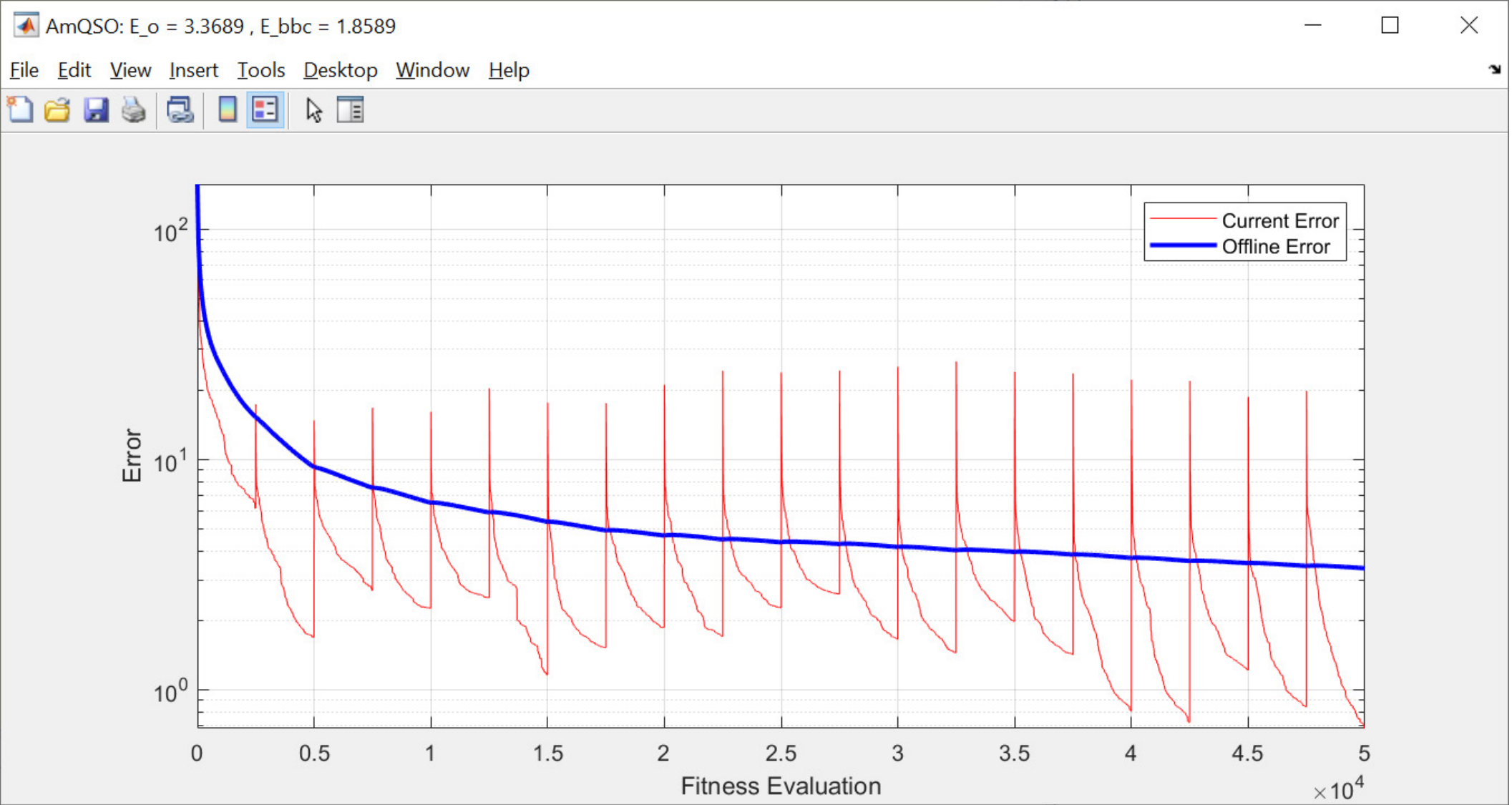}
\caption{An output figure of an experimentation in EDOLAB.
This figure depicts the plots of offline and current errors over time.
The plots are the average of all runs.}
\label{fig:OutputFig}
\end{figure}

To track progress, EDOLAB displays the current run number and environment in the MATLAB Command Window. 
{After the experiment is complete, the average, median, standard error values of the performance indicators, and runtime statistics are displayed in the Command Window.
The detailed results of individual runs, along with their averages, medians, standard error values, runtime data, and the main benchmark parameters, are saved in an Excel file located in the $\mathtt{Results}$ folder, if the corresponding checkbox was selected.}
The Excel file name includes the EDOA, benchmark name, and the date and time of the experiment (e.g., $\mathtt{EDOA\_Benchmark\_DateTime.xlsx}$). 
These results and statistics can be used for further statistical analysis using MATLAB or other software. 
An example of an Excel file generated by EDOLAB is shown in Figure~\ref{fig:Excel}. 
Additionally, if the user selected the relevant checkbox, a figure with plots of the offline and current errors over time is generated. 
An example of the output plots is provided in Figure~\ref{fig:OutputFig}.

\paragraph{{A Note on Parameter Settings of Benchmarks}}
\label{sec:sec:suggestedSettings}

The four benchmark generators included in EDOLAB share several common parameters, which have been widely manipulated in the literature to generate problem instances with varying levels of difficulty and characteristics. 
The GUI in EDOLAB facilitates the adjustment of these parameters, allowing users to easily configure benchmark scenarios. 
In Table~\ref{tab:BenchmarkParameters}, we provide suggested values for these parameters, which can be used to generate standardized problem instances for comparing the performance of different algorithms.
The key parameters that can be adjusted include:
\begin{itemize}
    \item {Number of Promising Regions}: Defines the number of promising regions in the search space.
    \item {Shift Severity}: Controls how significantly the search space changes between environments.
    \item {Dimension}: Sets the number of variables in the optimization problem.
    \item {Change Frequency}: Specifies how often the environment changes during the optimization process.
\end{itemize}

To create a well-rounded experimental setup, we recommend selecting one benchmark generator from FPs or MPB and one from GDBG or GMPB. This approach ensures a balance between simpler and more complex problem instances, allowing for a more comprehensive evaluation of algorithm performance. 
FPs and MPB represent benchmarks with fewer challenges, making them suitable for baseline comparisons, while GDBG and GMPB introduce more difficult problem instances with complex characteristics. 
This diversity helps to test the EDOAs' ability to adapt to varying levels of difficulty and complexity. 
For each chosen benchmark generator, apply the parameter settings provided in Table~\ref{tab:BenchmarkParameters}. 
By using the different parameter settings provided in Table~\ref{tab:BenchmarkParameters}, we can generate 12 distinct problem instances for each chosen benchmark generator.

Using the suggested parameter settings in Table~\ref{tab:BenchmarkParameters}, researchers can generate diverse problem instances from the included benchmark generators, helping to establish a standardized experimental setup for algorithm comparison. These settings provide a common foundation for most studies in the field of evolutionary dynamic optimization. However, it is important to consider that specific studies may require different parameter settings, depending on the scope and focus of the research. For example, higher dimension values are used in research focused on large-scale dynamic optimization~\cite{bai2022evolutionary} 
Ultimately, while these suggestions aim to provide a consistent framework for comparison, they can be adapted to suit the requirements of targeted studies.

\begin{table}[t]
\small

\centering
 \caption{Types and suggested values for the main parameters of the benchmark generators in EDOLAB. The highlighted values represent the default settings for each parameter. When testing algorithms on specific parameters (e.g., different dimensions), the other parameters should be set to their default values to generate consistent problem instances.}
 \label{tab:BenchmarkParameters}
 \begin{threeparttable}
    \begin{tabular}{llll}
    \toprule
    Parameter   & Name in the source code &Type & Suggested values\\
    \midrule 
Dimension & $\mathtt{Problem.Dimension}$& Positive integer &  $\in \{2,\hl{5},10,20\}$\tnote{\textbf{$\star$}}\\
Number of promising regions & $\mathtt{Problem.PeakNumber}$ & Positive integer &  $\in \{\hl{10}, 25,50,100\}$\\
Change frequency & $\mathtt{Problem.ChangeFrequency}$ & Positive integer &  $\in \{500,1000,2500,\hl{5000}\}$\\
Shift severity & $\mathtt{Problem.ShiftSeverity}$ & Non-negative real valued &  $\in \{\hl{1},2,5\}$\\
Number of environments   & $\mathtt{Problem.EnvironmentNumber}$ & Positive integer &  $100$\tnote{\textbf{$\dagger$}}\\
 \bottomrule
  \end{tabular}
    \begin{tablenotes}
   \begin{scriptsize}
  \item[$\star$] These are suggested values for the experimentation module.
  In the education module, the dimension can only be set to two.
  \item[$\dagger$] For the sake of understandability, the number of environments is suggested to set between 10 and 20 in the education module. 
   \end{scriptsize}
    \end{tablenotes}
 \end{threeparttable}
\end{table}

\color{black}

\subsubsection*{Education module}
\begin{figure}[!t]
\centering
\includegraphics[width=0.9\linewidth]{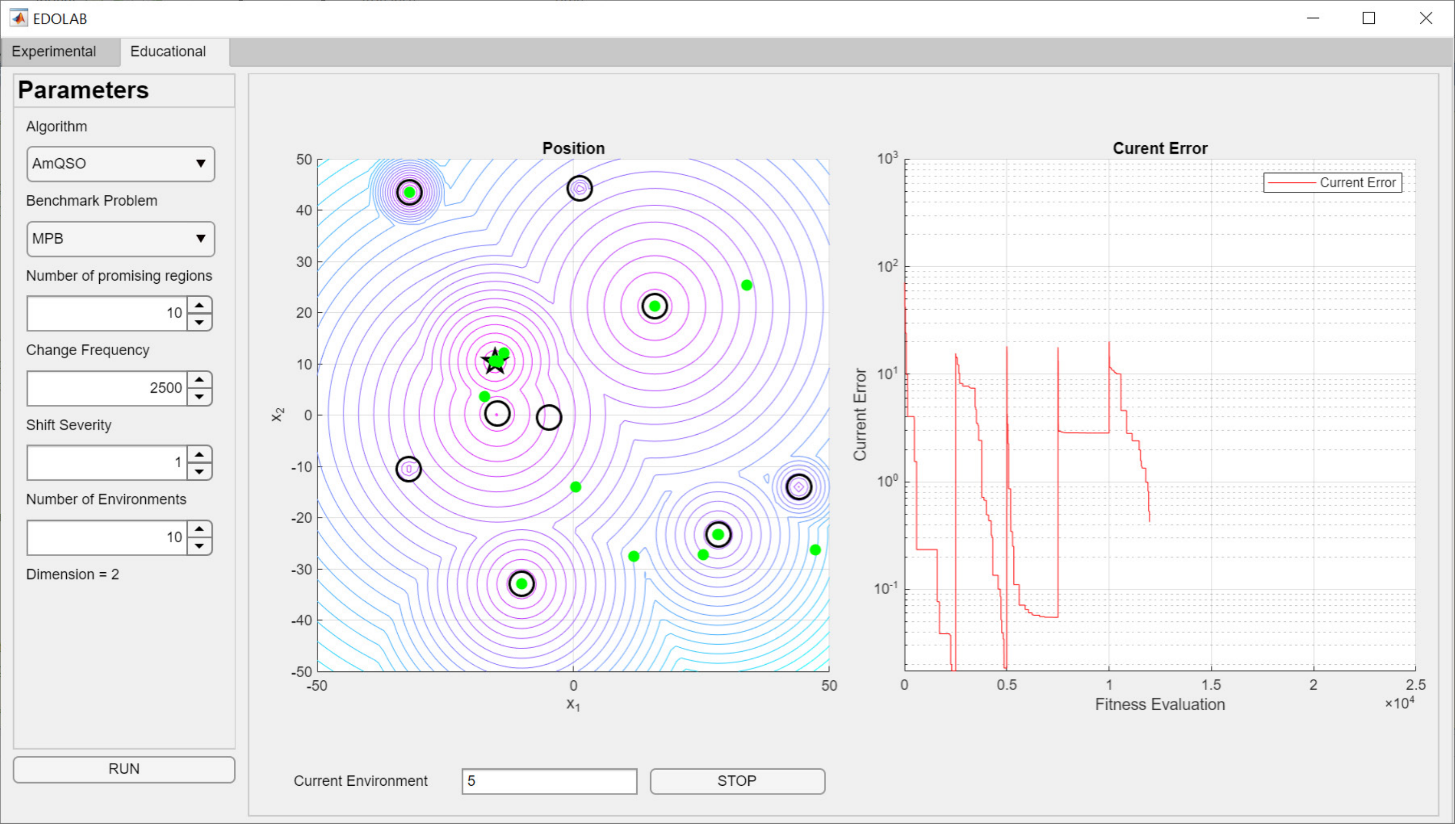}
\caption{The education module of EDOLAB.}
\label{fig:EducationModul}
\end{figure}

The education module allows users to visually observe the current environment, environmental changes, and the positions and behaviors of individuals over time. 
Figure~\ref{fig:EducationModul} displays the interface of EDOLAB's education module. 
On the left side of the interface, users can configure an experiment in a manner similar to the experimentation module.
However, only 2-dimensional problem instances are supported in the education module, as the goal is to visualize the problem space and individuals.

Once the experiment is configured and the $\mathtt{RUN}$ button is pressed, the experiment begins, and the environmental parameters and positions of individuals over time are archived. 
The time required for the run will depend on the CPU, the selected EDOA, and the benchmark settings. 
After the run is complete, the archived information is displayed within the education module interface.

Using the archived data, the education module generates a video showing the environments and the positions of individuals over time. 
As depicted in Figure~\ref{fig:EducationModul}, a 2-dimensional contour plot is used for this visualization. 
In the contour plot, the center of each visible promising region---those not covered by larger regions---is marked with a black circle, the global optimum position is indicated by a black pentagram, and the individuals are represented by green filled circles. 
The positions of individuals are updated every iteration, and the contour plot is refreshed after each environmental change.
Additionally, the current error plot and the current environment number are shown to provide further insights, which enhance the understanding of the problem and the behavior of the EDOA.

By monitoring the positions of individuals, the search space/environment, the current environment number, and the current error plot over time, users can observe the EDOA's performance in exploration, exploitation, and tracking within each environment. 
Furthermore, the effectiveness of various EDOA components---such as mutual exclusion in promising regions~\cite{blackwell2006multiswarm}, the generation of new sub-populations~\cite{blackwell2008particle}, promising region coverage, mechanisms for increasing global diversity, and change reaction---can also be analyzed using the education module.

Unlike the experimentation module, where identical random streams are used across all experiments, the education module employs different random streams for each run. 
Consequently, users can observe the behavior and performance of the EDOA in different problem instances during each run of the education module.

\subsection{Using EDOLAB without GUI}
\label{sec:sec:RunWithoutGUI}
EDOLAB can also be operated without the GUI, which offers more advanced and flexible options for users. 
In this mode, users interact directly with the source codes of EDOLAB, which enables them to: 
\begin{inparaenum} 
\item modify the parameter settings of the EDOAs, 
\item alter or disable certain components of the EDOAs, and 
\item adjust all parameters of the benchmarks. 
\end{inparaenum} 
To enhance readability, understanding, and ease of navigation within EDOLAB's source code, we have: 
\begin{itemize} 
\item divided the code into \emph{sections} using the $\%\%$ command, with each section having a descriptive header. 
These sections group related lines of code, such as those implementing components (for example, exclusion~\cite{blackwell2006multiswarm}), initializing EDOA parameters, or preparing output values, 
\item assigned meaningful and descriptive names to all structures, parameters, and functions in EDOLAB, and 
\item added informative comments throughout the code to assist users.  
\end{itemize}

To run an EDOA without the GUI, users interact with the $\mathtt{RunWithoutGUI.m}$ file in the root directory of EDOLAB.
Within this file, users can select the EDOA, choose the benchmark, and configure the main benchmark parameters (as shown in Table~\ref{tab:BenchmarkParameters}). 
To specify an EDOA and benchmark, the user sets $\mathtt{AlgorithmName}$ to the desired EDOA (for example, $\mathtt{AlgorithmName}\;=\;'\mathtt{mQSO}^\backprime$) and $\mathtt{BenchmarkName}$ to the desired benchmark (for example, $\mathtt{BenchmarkName}\;=\;'\mathtt{GMPB}^\backprime$).

Users can also choose between the experimentation and education modules within $\mathtt{RunWithoutGUI.m}$. 
Similar to the GUI's education module (see Figure~\ref{fig:EducationModul}), selecting the education module in $\mathtt{RunWithoutGUI.m}$ will display contour plots of the environments, the positions of individuals, and the current error over time. 
The education module is activated when the user sets $\mathtt{VisualizationOverOptimization=1}$.

If $\mathtt{VisualizationOverOptimization=0}$ is set, the experimentation module is activated. 
When using the experimentation module, users can configure the outputs. 
By setting $\mathtt{OutputFigure}$ to 1, users can generate visual plots of offline and current errors (see Figure~\ref{fig:OutputFig}). 
Additionally, setting $\mathtt{GeneratingExcelFile=1}$ will save an Excel file containing output statistics and results in the $\mathtt{Results}$ folder. 
These archived results in the Excel file can later be used for statistical analysis.

The parameters of the selected EDOA can also be modified in its main function (for example, $\mathtt{main_mQSO.m}$), which is located in the EDOA's sub-folder. 
By default, these parameters are set to the values recommended in their original references. 
The lines of code for initializing EDOA parameters are found in the $\mathtt{\%\%~Initializing~Optimizer}$ section of the EDOA's main function. 
A structure named $\mathtt{Optimizer}$ contains all the parameters of the EDOA.

In addition to the main parameters of the benchmark generators listed in Table~\ref{tab:BenchmarkParameters}, each benchmark has additional parameters. 
Typically, researchers modify only the main parameters to generate different problem instances. 
However, users wishing to evaluate EDOA performance on instances with specific characteristics can adjust other parameter values in the corresponding $\mathtt{BenchmarkGenerator\_Benchmark.m}$ file. 
For example, Table~\ref{tab:GMPBsettings} shows the parameters of GMPB that can be altered by the user in $\mathtt{BenchmarkGenerator\_GMPB.m}$, located in $\mathtt{EDOLAB\backslash Benchmark\backslash GMPB}$.

\begin{table}[tp!] 
    \small
\centering
  \caption{Parameters of GMPB that can be changed by the user to generate problem instances with different morphological and dynamical characteristics. } 
  \label{tab:GMPBsettings}
 \begin{threeparttable}
  \begin{tabular}{llll}
    \toprule
    Parameter & Name in the source code & Suggested value(s)\\ 
\midrule
Dimension\tnote{\textbf{$\dagger$}}   & $\mathtt{Problem.Dimension}$	&   $\in \{1,2,5,10\}$\\
Numbers of promising regions\tnote{\textbf{$\dagger$}}  & $\mathtt{Problem.PeakNumber}$	             & $\in \{10, 25,50,100\}$\\
Change frequency\tnote{\textbf{$\dagger$}}   & $\mathtt{Problem.ChangeFrequency}$	             & $\in \{500,1000,2500,5000\}$\\
Shift severity\tnote{\textbf{$\dagger$}}   & $\mathtt{Problem.ShiftSeverity}$	&    $\in \{1,2,5\}$\\
Number of environments\tnote{\textbf{$\dagger$}}     &   $\mathtt{Problem.EnvironmentNumber}$       &  $100$ \\
\midrule
Height severity   & $\mathtt{Problem.HeightSeverity}$	    & 7 \\
Width severity    & $\mathtt{Problem.WidthSeverity}$	    & 1\\
Irregularity parameter $\tau$ severity    & $\mathtt{Problem.TauSeverity}$	    &  0.2 \\
Irregularity parameter $\eta$ severity    & $\mathtt{Problem.EtaSeverity}$	    &  10 \\
Angle severity    & $\mathtt{Problem.AngleSeverity}$	    &  $\pi / 9$ \\
Search range upper bound                        &   $\mathtt{Problem.MaxCoordinate}$	    &    $50$ \\
Search range lower bound                       &   $\mathtt{Problem.MinCoordinate}$	    &    $-50$ \\
Maximum height                         &  $\mathtt{Problem.MaxHeight}$  	    &    $70$   \\
Minimum height                         &  $\mathtt{Problem.MinHeight}$  	    &    $30$   \\
Maximum width                          &   $\mathtt{Problem.MaxWidth}$  	    &   $12$  \\
Minimum width                          &   $\mathtt{Problem.MinWidth}$  	    &   $1$  \\
Maximum angle                           & $\mathtt{Problem.MaxAngle}$ & $\pi$  \\
Minimum angle                           & $\mathtt{Problem.MinAngle }$ & $-\pi$  \\
Maximum irregularity parameter $\tau$   &   $\mathtt{Problem.MaxTau}$ &$1$ \\
Minimum irregularity parameter $\tau$   &   $\mathtt{Problem.MinTau}$ &$0.1$ \\
Maximum irregularity parameter $\eta$    &  $\mathtt{Problem.MaxEta}$ & $50$  \\
Minimum irregularity parameter $\eta$   &  $\mathtt{Problem.MinEta}$ & $0$  \\
    \bottomrule
  \end{tabular}
  \begin{tablenotes}
   \begin{scriptsize}
  \item[$\dagger$] These are commonly used parameters to generate different problem instances with various characteristics.
  As stated before, these parameters are common among the benchmark generators of EDOLAB and can be either set in the GUI or $\mathtt{RunWithoutGUI.m}$. 
    \end{scriptsize}
    \end{tablenotes}
 \end{threeparttable}
 \end{table}

Once the configurations are complete, the user can run $\mathtt{RunWithoutGUI.m}$ to initiate the experiment. 
During the run, progress information---including the current run number and environment number---is displayed in the MATLAB Command Window. 
Upon completion of the experiment, the results are also presented in the MATLAB Command Window.

\section{Extension}
\label{sec:sec:extension}

Users can extend EDOLAB, as it is an open-source platform. 
Below, we describe how to add new benchmark generators, performance indicators, and EDOAs to EDOLAB.

\subsection{Adding a benchmark generator}

Suppose a user wants to add a new benchmark called ABC. 
First, the user must create a new sub-folder named ABC within the $\mathtt{Benchmark}$ folder. 
Then, two functions, $\mathtt{fitness\_ABC.m}$ and $\mathtt{BenchmarkGenerator\_ABC.m}$, need to be added to this folder.

In $\mathtt{BenchmarkGenerator\_ABC.m}$, the user defines and initializes all the parameters of the new benchmark within a structure named $\mathtt{Problem}$, similar to how the parameters of existing benchmark generators in EDOLAB are defined. 	
Subsequently, the environmental parameters for all environments must be generated in this function, and all the environmental and control parameters of ABC must be stored in the $\mathtt{Problem}$ structure.

The second function, $\mathtt{fitness\_ABC.m}$, contains the code for the baseline function of ABC. 
Both $\mathtt{BenchmarkGenerator\_ABC.m}$ and $\mathtt{fitness\_ABC.m}$ must have inputs and outputs consistent with those of EDOLAB's current benchmarks. 
No changes are required in other functions, and ABC will automatically be added to the list of benchmarks in the GUI and can also be accessed via $\mathtt{RunWithoutGUI.m}$.

\subsection{Adding a performance indicator}

Typically, the information required for calculating performance indicators in dynamic optimization problem (DOP) literature is gathered over time---either at the end of each environment~\cite{trojanowski1999searching}, after every function evaluation~\cite{branke2003designing}, or when solutions are deployed in each environment~\cite{yazdani2018thesis}. 
In EDOLAB, this data is collected in $\mathtt{fitness.m}$ and stored in the $\mathtt{Problem}$ structure.

To add a new performance indicator, the user first needs to modify $\mathtt{fitness.m}$ to collect the necessary data and store it in the $\mathtt{Problem}$ structure. 
The code for calculating the performance indicator should then be added to the $\mathtt{\%\%~Performance~indicator~calculation}$ section in the main function of the EDOA (e.g., $\mathtt{main\_mQSO.m}$).
Additionally, the results of the newly added performance indicator must be included in the outputs, which can be done in the $\mathtt{\%\%~Output~preparation}$ section at the bottom of the EDOA's main function.

\subsection{Adding an EDOA}

Adding a new EDOA to EDOLAB requires minimal modifications to the source code to ensure compatibility. 
Users should follow these steps:

\begin{itemize}
    \item First, create a sub-folder inside the $\mathtt{Algorithm}$ folder, named according to the new EDOA. 
    Then, add the EDOA's functions to this sub-folder.
    
    \item The new EDOA must be invoked by $\mathtt{RunWithoutGUI.m}$. 
    The user should ensure that the inputs and outputs of the EDOA's main function are compatible with $\mathtt{RunWithoutGUI.m}$.
    
    \item In the main function of the new EDOA, call $\mathtt{BenchmarkGenerator.m}$ to generate the problem instance.
    
    \item To enable the education module, include the code that generates and collects information related to the education module in the main loop of the EDOA. 
    This code can be found in the $\mathtt{\%\%~Visualization~for~education~module}$ section of other EDOAs.
    
    \item Use $\mathtt{fitness.m}$ for evaluating the fitness of solutions.
    
    \item Before initializing the optimizer in the main function of the EDOA, define parameters and data structures for gathering runtime, performance indicators, and other output information. After each run, ensure that the necessary information is stored in these parameters and arrays.
    
        \item Before initializing the optimizer in the main function of the EDOA, define parameters and data structures for gathering runtime, performance indicators, and other output information. After each run, ensure that the necessary information is stored in these parameters and arrays.

        \item At the end of the main function of the EDOA, include the code for output preparation similar to the structure in the existing algorithms.
    
    \item The main function of the newly added EDOA should be named $\mathtt{main\_EDOA.m}$ to make it accessible through EDOLAB.
\end{itemize}

For example, if the new EDOA is called XYZ, the sub-folder should be named XYZ, and the main function file should be named $\mathtt{main\_XYZ.m}$. 
Once this is done, the new EDOA will automatically be added to the list of available algorithms in both the GUI modules. Additionally, by setting $\mathtt{AlgorithmName}\;=\;'\mathtt{XYZ}^\backprime$, the algorithm can be run using $\mathtt{RunWithoutGUI.m}$.

\section{Using EDOLAB in Octave}
\label{sec:Octave}

Since EDOLAB was originally developed in MATLAB, some users may prefer to use an open-source alternative to run the platform. Octave is a widely-used open-source software that is largely compatible with MATLAB, providing researchers with a free alternative to access EDOLAB's features without requiring a MATLAB license. With minor modifications, many of EDOLAB's functionalities can be used in Octave, though there are certain limitations.
One major limitation is that the GUI functionality is not supported in Octave, as it relies on MATLAB's App Designer. Consequently, all experiments and tasks in Octave must be carried out through \texttt{RunWithoutGUI.m}.
Below are guidelines on how to use EDOLAB with Octave and the necessary modifications to ensure compatibility.

\subsection{Notes on Using \texttt{RunWithoutGUI} in Octave}

While Octave is largely compatible with MATLAB, there are a few important differences to keep in mind when running \texttt{RunWithoutGUI.m} in Octave.

\begin{itemize}
    \item \textbf{Necessary packages}: 
    The \texttt{statistics} and \texttt{io} packages must be loaded to run EDOLAB in Octave. These packages provide functions essential for statistical computations, distance calculation (\texttt{pdist2} function), and reading/writing files.

    \item \textbf{Generating Excel Files}: 
    \begin{itemize}
        \item The \texttt{xlswrite} function requires the \texttt{io} package, which is automatically loaded.
        \item ActiveX is not supported in Octave, meaning Excel files cannot be opened automatically after they are generated.
        \item The \texttt{actxserver} function, which MATLAB uses to control Excel, is unavailable in Octave.
        \item Font color, font style, and other formatting options are unsupported, so Excel files generated in Octave will have a basic, unformatted style.
    \end{itemize}

    \item \textbf{Generating Plots}: 
    \begin{itemize}
        \item Functions like \texttt{append}, \texttt{parfor}, and \texttt{blkdiag} (used within the append function) are not supported in Octave. These are needed for advanced plotting, so alternative methods may be required to replicate this functionality.
    \end{itemize}

    \item \textbf{Calling and Writing Benchmark and Algorithm Names}:
    \begin{itemize}
        \item Octave does not support the \texttt{""} syntax for strings. Instead, users must use cell arrays for strings within the \texttt{RunWithoutGUI.m} script.
    \end{itemize}


    \item \textbf{Free Peaks (FPs) Benchmark}:
    \begin{itemize}
        \item Octave cannot read \texttt{.mexw64} files generated by MATLAB from \texttt{.cpp} files. To resolve this, delete the existing \texttt{.mexw64} file in the KDTree folder and generate a new \texttt{.mex} file using the following command:
        \begin{verbatim}
        mkoctfile -v --mex ConstructKDTree.cpp
        \end{verbatim}
        \item Ensure that Octave is in the same directory as the \texttt{.cpp} file when running this command.
        \item If errors occur, you may need to install the MinGW-w64 compiler.
    \end{itemize}
\end{itemize}

\subsection{Octave Compatibility Folder}

To simplify the process of using EDOLAB in Octave, we have created a folder named \texttt{Octave\_compatibility}, which contains all the files that have been updated for compatibility with Octave. These modifications ensure that the core functionalities of EDOLAB work without issues in Octave. The key changes include:

\begin{itemize}
    \item \texttt{RunWithoutGUI}: Adapted to work with Octave's syntax and configured to automatically load the necessary packages (\texttt{statistics} and \texttt{io}).
    \item \texttt{OutputExcel}: Simplified to function without ActiveX or advanced Excel formatting features.
    \item \texttt{OutputPlot}: Adjusted to account for limitations in Octave's plotting capabilities.
    \item \texttt{KDTree}: The generation of \texttt{.mex} files from \texttt{.cpp} files must be performed manually, as described earlier.
\end{itemize}

Users wishing to run EDOLAB in Octave can do so by replacing the corresponding files in the main EDOLAB directory with the modified files provided in the \texttt{Octave\_compatibility} folder. Once the files are replaced, \texttt{RunWithoutGUI.m} can be executed in Octave to run experiments and tasks without further adjustments.

\color{black}

\bibliography{bib}
\bibliographystyle{ACM-Reference-Format}

\end{document}